\pdfoutput=1
\documentclass[11pt]{article}

\usepackage{EMNLP2023}
\usepackage{times}
\usepackage{latexsym}

\usepackage[T1]{fontenc}
\usepackage[utf8]{inputenc}

\usepackage{microtype}

\usepackage{inconsolata}
\usepackage{adjustbox}
\usepackage{amsmath}
\usepackage{amsfonts}
\usepackage{graphicx}
\usepackage{subcaption}
\usepackage{booktabs}
\usepackage{cleveref}

\usepackage{breqn}
\title{Learning to Abstract with \\ Nonparametric Variational Information Bottleneck}
\author{Melika Behjati$^{\dag ~1,2}$ \\
  \And
  Fabio Fehr$^{\dag ~1,2}$ \\
  $^1$ Idiap Research Institute, Switzerland \\
  $^2$ École Polytechnique Fédérale de Lausanne, Switzerland \\
  \texttt{firstname.lastname@idiap.ch} \\
  \And
  James Henderson$^{~1}$ \\
  }

\newcommand{\customfootnotetext}[2]{{% Group to localize change to footnote
  \renewcommand{\thefootnote}{#1}% Update footnote counter representation
  \footnotetext[0]{#2}}}% Print footnote text

\begin{document}
\maketitle

\customfootnotetext{\dag}{Equal contribution.}

\begin{abstract}
% Introducing levels of abstraction
Learned representations at the level of characters, sub-words, words and sentences, have each contributed to advances in understanding different NLP tasks and linguistic phenomena.
% Problem
However, learning textual embeddings is costly as they are tokenization specific and require different models to be trained for each level of abstraction.
% Model
We introduce a novel language representation model which can learn to compress to different levels of abstraction at different layers of the same model. 
% How it works
We apply Nonparametric Variational Information Bottleneck (NVIB) to stacked Transformer self-attention layers in the encoder, which encourages an information-theoretic compression of the representations through the model. 
% Results + claims 
% Visual + arxiv abstraction claim
We find that the layers within the model correspond to increasing levels of abstraction
% Linguistically informed (SentEval) claim: (HARD to show)
and that their representations are more linguistically informed.
% Robustness claim:
Finally, we show that NVIB compression results in a model which is more robust to adversarial perturbations. 
\end{abstract}

%\fabio{More linguistically informed. Not different phenomena}

\section{Introduction}

Learning representations of language using self-supervision has become a cornerstone of NLP \citep[\textit{inter alia}]{pennington-etal-2014-glove,peters-etal-2018-deep, devlin-etal-2019-bert}.  However, these representations are specific to their tokenisation (e.g.\ Byte-Pair \cite{sennrich-etal-2016-neural}, WordPiece \cite{schuster2012japanese}, SentencePiece \cite{kudo-richardson-2018-sentencepiece}, characters \cite{al2019character}, and even bytes \cite{xue-etal-2022-byt5}), which restricts the level of abstraction from the input text which their representations are able to convey.  Work like CANINE \cite{clark-etal-2022-canine} and Charformer \cite{tay2022charformer} avoid problems with tokenisation by modeling individual characters or bytes, and thereafter use a stride-based downsampling to reduce the representation length.  The stride pattern is fixed and thus can't be considered as learning to abstract.  \citet{behjati2023inducing} recently introduced the task of learning a higher level of abstraction in a set-of-vector space by proposing Dynamic Capacity Slot Attention. 
In this work, we propose a novel character-level model of representation learning which learns different levels of abstraction in different layers of the same model.  

\begin{figure}[t]
    \centering
    \includegraphics[width=0.25\textwidth]{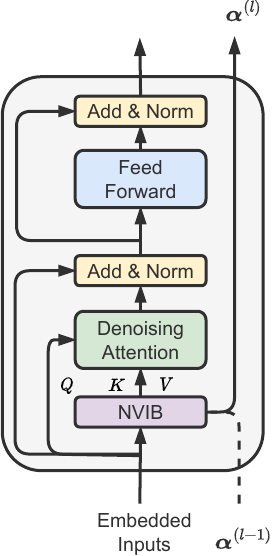}
    \caption{
    % \textbf{Left}: 
    Transformer encoder layer $(l)$ including the NVIB layer and Denoising self-attention module. 
    % \textbf{Right}: Illustrative
    % Expected 
    % attention patterns with abstraction at higher levels.
    }
    \label{fig:model_figure}
\end{figure}
% \fabio{Make smaller? Crop out white space. Inline?}

\paragraph{Contributions}
%More specifically, 
We adapt the Nonparametric Variational Information Bottleneck regulariser (NVIB) \cite{hendersonfehr2023} for application to self-attention in the stacked layers of a Transformer encoder.\footnote{
The code is publically available at: \\ 
% Package: 
\url{https://github.com/idiap/nvib} \\ 
% Paper: 
\url{https://github.com/idiap/nvib_selfattention}} The resulting model has greater abstraction than a standard Transformer due to selectively dropping some vectors in higher attention layers.
Interestingly, we observe that the learned abstract units are intuitive, often corresponding to words.  
% Our model is also more robust to noise and better at encoding semantically and linguistically meaningful information.
By employing different analysis methods, we demonstrate that our model is 
% always
better at encoding semantically and linguistically meaningful information than a standard Transformer baseline. Moreover, it exhibits an enhanced level of robustness, further consolidating its advantage.

\section{The Model}
Our model consists of standard Transformer encoder-decoder layers \cite{vaswani2017}, where the encoder block has been augmented with an NVIB regulariser on the self-attention layers, as seen in Figure \ref{fig:model_figure}.  
% NVIB induces 
%sparsity, in that 
% some of the vectors in the latent representation to be dropped, shown as blank columns on the right.
%This requires three components: the NVIB layer which outputs a sample from the distribution; the denoising self-attention function which interprets the output as a mixture distribution; and the VIB loss which regularises the distribution parameters.  

\subsection{NVIB for Self-Attention} %Nonparametric Variational Information Bottleneck

% Background to previous NVIB
%Nonparametric Variational Information Bottleneck 
Nonparametric Variational Information Bottleneck is an information-theoretic regulariser for attention-based latent representations \cite{hendersonfehr2023}. It has been shown to induce smooth and sparse latent representations in the cross-attention layer of a Transformer encoder-decoder, where \citet{hendersonfehr2023} used it to define a Variational Auto-Encoder (VAE) \citep{kingma2014autoencoding}.  It generalises attention over a set of vectors to \textit{denoising attention} over a mixture of impulse distributions, and uses Bayesian nonparametrics to handle the fact that the number of vectors grows with the length of the text.  NVIB uses Dirichlet Processes (DPs) to define distributions over these mixture distributions, and controls the information in the latent representation by sampling a mixture distribution from the attention layer's DP, thereby adding noise which removes information.   

We extend the previous work by using implicit reparameterisation gradients \cite{Figurnov2018ImplicitRG} to improve learning, and by adapting NVIB for use in the stacked self-attention layers of a Transformer encoder. 
By extending NVIB's information-theoretic regularisation to the series of latent representations inside the Transformer encoder, we see increasingly abstract interpretable representations in the higher layers. 

\paragraph{NVIB layer} 
% Old NVIB definition
As with a standard attention layer, an NVIB layer maps a set of $n$ vectors to an attention function.  It first maps the $n$ vectors $\boldsymbol{Z} \in \mathbb{R}^{n \times p}$ to the parameters of a DP, which are a total pseudo-count for its Dirichlet distribution and a mixture of Gaussians for its base distribution.
% For each of the individual $n$ tokens, the layer linearly projects the representation to three parameters: ${\alpha}_i \in \mathbb{R}$, $\boldsymbol{\mu}_i \in \mathbb{R}^{1 \times p} $ and $\text{log}(\boldsymbol{\sigma}_i) \in \mathbb{R}^{1 \times p}$, where the variance parameters are exponentiated to be strictly positive. Our implementation  replaces the previous activation (ReLU) for the pseudo-count parameters ${\alpha}_i$ with an exponential activation and provide a skip connection from the previous layers:
% \begin{align}
%     {{\alpha}_i}^{(l)} &= \text{exp}(\boldsymbol{w} \boldsymbol{z}_i + b + \text{log}({{\alpha}_i}^{(l-1)})),
% \end{align}
Each of the $n$ vectors is individually projected to a pseudo-count $\boldsymbol{\alpha} \in \mathbb{R}^n$ and a Gaussian component $(\boldsymbol{\mu} \in \mathbb{R}^{n \times p},\boldsymbol{\sigma} \in \mathbb{R}^{n \times p})$ of the base distribution.  
The model can drop entire vectors by setting their pseudo-counts to zero, thereby making the representation sparse.
In addition, there is an $n{+}1^{th}$ component of the base distribution for the prior, with parameters $\alpha^p{=}1$, $\boldsymbol{\mu}^p{=}\boldsymbol{0}$ and $\boldsymbol{\sigma}^p{=}\boldsymbol{1}$.
The individual pseudo-counts are both summed to get the DP's total pseudo-count and normalised to weight the components of the DP's base distribution.
The NVIB layer then uses denoising attention to access either a set of weighted vectors sampled from the DP (at training time), or the base distribution of the DP (at testing time).

\citet{hendersonfehr2023} use ReLU, linear and exponential activation functions to compute $\boldsymbol{\alpha}$, $\boldsymbol{\mu}$ and $\boldsymbol{\sigma}$, respectively.
% Our contribution 
To adapt NVIB for stacked layers of self-attention, our model replaces the activation for the pseudo-count parameters with an exponential activation, and includes a multiplicative skip connection from the previous layer $l{-}1$, as shown in Figure \ref{fig:model_figure}:\nolinebreak
% \vspace{-1ex}
\begin{align}
    {\boldsymbol{\alpha}}^{(l)} &= \text{exp}(\boldsymbol{w} \boldsymbol{Z}^T + b + \text{log}({\boldsymbol{\alpha}}^{(l-1)})),
  % \\
  % [-5ex]~\nonumber
\end{align}
where $\boldsymbol{w} \in \mathbb{R}^{1 \times p}$ and $b \in \mathbb{R}$ form the linear projection. The exponential activation allows the model to be more stable in training.\footnote{Since the exponential function is never exactly zero, we threshold small values to introduce sparsity. See Appendix~\ref{apx:training}.} The skip connection in between layers $l{-}1$ and $l$ helps coordinate the importance of vectors across layers. Keeping the pseudo-count parameters in log-space prevents overflow and improves precision when the parameters get larger. This results in a multiplicative  skip connection which emphasizes the communication between layers. 

% Overflow The pseudo-count parameters 
%  control the noise in sampling during training and bias the attention weights. During initial training steps with KL annealing the model is incentivised to make the 
%  larger. This is due to low initial regularisation in KL annealing. For overflow and precision the original NVIB implementation kept the 
%  in log space, thus we thought it would be better to keep the skip connection in log space.
% Magnitude The skip connection was put in place to allow communication in attention between layers and guide the abstraction. Since we considered only 3 layers of NVIB a multiplicative skip connection has a greater magnitude than an additive skip connection. Considering an additive skip connection is a valuable suggestion and we suspect that this may be helpful when considering many more connected NVIB layers.

To compute self-attention, the DP parameters projected from all the individual vectors together define a single DP, and we take a single sample from this DP which all the individual vectors access via denoising attention.  The queries for this denoising self-attention are computed from the original $n$ vectors $\boldsymbol{Z} \in \mathbb{R}^{n \times p}$, before the NVIB layer.  
%Backprop through this sampling step is done using the reparameterisation trick for the Gaussian components \cite{kingma2014autoencoding} and 
We also introduce the use of implicit reparameterisation gradients \cite{Figurnov2018ImplicitRG} for 
%the Dirichlet distributions
error backpropagation through the sampling step.
See Appendix \ref{apx:dattn} for the exact attention functions.

\paragraph{Training objective}
% Describe the NVIB Loss function
The NVIB loss regularises the attention-based representations so that the size of the representation at each layer is appropriate for the complexity of the representation being encoded at that layer. 
%This is done by applying evidence lower bound (ELBO) objective which approximately maximises the log-likelihood of the observation at each layer. This loss can be constructed as 
It has three terms, a reconstruction loss $L_R$, and two KL divergence 
%loss between each layer's posterior and the prior distribution. The KL divergence has 
terms: $L_D$ for the pseudo-counts of the Dirichlet distributions, and $L_G$ for the parameters of the Gaussian components.  
The $L_R$ term is the supervised learning objective, which tries to make the latent representation informative enough to predict the original text.
The $L_G$ term tries to make the individual Gaussian components less informative, as in vector-space VAEs \citep{kingma2014autoencoding}.
The $L_D$ term tries to push down the total pseudo-count, which pushes some of the individual pseudo-counts to zero, thereby effectively dropping their vectors and reducing the number of vectors in the latent representation.
See Appendix \ref{apx: kl_loss} for loss equations.
%for $L_D$ and $L_G$.

% Describe the addition we have made
To apply NVIB to stacked self-attention layers, we want to allow the lower layers to compute with more vectors while encouraging the upper layers to compress to fewer vectors, thereby encouraging abstraction at the higher layers. We therefore weight the loss terms differently at each layer:
%Each layer of NVIB self-attention needs to be regularised. To coax the sequential abstraction through the layers we define a layer-wise weighted loss:  
%
% es. During initial exploration of the regularisation we found a uniform regularisation strength across layers did not lead to the same abstraction. Often the lower levels were penalised too strongly. We also considered a regularisation that would double in strength through the layers, which is more in line with the multiplicative skip connection. This regularisation was too strong and lead to the final layer being compressed to a single sentence representation
%
% \vspace{-1ex}
\begin{align}
    \label{eq:loss}
  \mathcal{L} &= L_R  ~+~ \beta^{(l)}(\lambda_D L_D  ~+~ \lambda_G L_G) \\
  \label{eq:loss_weighting}
  \beta^{(l)} &= \frac{l}{\sum_{l=0}^N l} \text{~~~~ for } l \in \{1, ..., N \}
  %   \beta^{(l)} &= \frac{2^l}{\sum_{l=0}^N 2^l} \text{ for } l \in \{0, ..., N \}
  % \\[-5ex]~\nonumber
\end{align}
where $\beta^{(l)}$ controls the degree of NVIB regularisation for layer $l$, linearly increasing it for higher layers.   If a vector is dropped in the last self-attention layer (i.e.\ zero pseudo-count), then we also drop that vector in the cross-attention layer to the decoder, but otherwise there is no NVIB regularisation of the cross-attention.

During preliminary experiments, instead of the above formula for $\beta^{(l)}$ we considered a uniform weight, as well as a doubling weight, per layer. These regularisation weights were either too weak or too strong, respectively. 
The values we considered for the hyperparameter $\lambda_D$ are given in Appendix~\ref{apx:tuning}.
When we increase this regularisation, the characters are grouped into fewer and fewer vectors until all characters are compressed into a single vector, much like a sentence embedding. 
% However, a single vector is challenging to get competitive reconstruction cross-entropy during validation. 
If we over-regularise, the representations collapse to the uninformative prior representation.

\section{Related Work}
Modeling language at the level of characters has the advantage of providing an end-to-end framework for the models to operate, without the need for tokenization as a preprocessing step \cite{xue-etal-2022-byt5,ataman2019latent,choe2019bridging,al2019character,kawakami-etal-2017-learning}. This is at the cost of longer sequence lengths and the need for greater model depth to reach the understanding level of subword-based models. While CANINE \cite{clark-etal-2022-canine} and Charformer \cite{tay2022charformer}  are some attempts to bypass these shortcomings, they do so by fixed architectural design choices. Our work differs in that it allows the model to learn how to abstract and compress the input without a hard-coded abstraction structure. 
% Recently, \citet{behjati2023inducing} introduced the task of learning a higher level of abstraction and proposed a method based on Slot Attention \cite{locatello2020object} for this purpose. 
Our inspiration comes from  \citet{behjati2023inducing} who introduced the task of learning a higher level of abstraction and proposed a method based on Slot Attention \cite{locatello2020object} for this purpose. 
%, through its NVIB-integrated layers without any explicit bias. 
Our work is also related to HM-RNNs \cite{chung2017hierarchical} as it tends to learn a hierarchy of units within its layers, though it does not make discrete decisions on unit boundaries.  Our approach to learning meaningful disentangled abstractions by encouraging the models to learn compressed representations through a bottleneck is shared with VAEs \cite{kingma2014autoencoding} and other work in that line \cite{alemi2017deep,higgins2017betavae}. 
% \paragraph{Analysis of NLP models.} Interpreting and analyzing how black-box NLP models make their decisions and what they learn from language has been an interest to the community in recent years (see \citet{belinkov-2019-analysis, madsen2022post} for a review). Therefore, different approaches have been proposed to unveil these question. Among them, probing has gained a lot of attention \cite{belinkov-2022-probing, pimentel-etal-2022-attentional, pimentel-etal-2020-information, hewitt-liang-2019-designing, hewitt-etal-2021-conditional, voita-titov-2020-information}. In addition, looking into the attention weights have revealed interesting patterns \cite{clark-etal-2019-bert}. These two approaches complement each other in providing an overall picture of what the model has learned. Orthogonal to these, analyzing the robustness of deep NLP models and discovering under which conditions the models fail has become important in building reliable models \cite{}. 

\section{Experiments}
% Analysis of abstraction.
% Analysis of the representations learned
% Evaluation of robustness
% Evaluation on a Downstream task

% Abstractness (qual -plots) - quantitave (arxiv)
% What are each layer learning (senteval)
% Robustness

Our proposed model's abstractness is analyzed qualitatively through attention visualisations (Section \ref{sec:attention}) and quantitatively through a challenging sub-topic classification task (Section \ref{sec:classification}). Each layer is probed to analyse the linguistic information captured (Section \ref{sec:probing}) and finally we examine the models' robustness to adversarial, synthetic noise (Section \ref{sec:robustness}). We provide additional details of these experiments in the \Cref{apx: probingclassifiers,apx:senteval,apx:f1,apx:arxiv}.

\subsection{Experimental Setup}

\paragraph{Data}
We train all models on the Wikitext-2 \cite{Merity2017PointerSM} encyclopedia dataset 
%was selected as it is a general English language corpora  containing high quality Wikipedia articles. We train the model 
at the character level, with a noisy character deletion reconstruction objective \cite{lewis-etal-2020-bart}.

\paragraph{Models}

We compare the self-attention representations from a standard Transformer encoder layer and our Transformer encoder layer with NVIB regularisation. We consider models consisting of six stacked Transformer encoder layers to be in line with the base model from \citet{vaswani2017}. For the Transformer decoder we use only 2 layers so that the decoder is not able to compensate for poor embeddings from the encoder. For simplicity of implementation and interpretation, we use only a single attention head. For the NVIB models, we only apply NVIB to the final three layers. To ensure comparability between our model and the baseline, we train the baseline to have the same denoising capability and thus the same validation cross-entropy when evaluated on noised examples. For further details see \Cref{apx:training,apx:tuning}. 
% For further training details see Appendix \ref{apx:training}. For hyperparameter tuning see Appendix \ref{apx:tuning} and final model hyperparameters see Table \ref{tab:Hyperparms}.
% \fabio{eThe current baselines have comparable validation CE when noising the input. Ie the same denoising ability and not necessarily the same reconstruction ability.}

% \subsection{Reconstruction Quality}
% We can have a comparison on the reconstruction quality of the two models. See \Cref{tab:reconstruction}.
% \melika{The NVIB works worse than a normal Transformer in terms of reconstruction loss and self-bleu, maybe we should remove this section.}

% \fabio{I think we can show this indirectly in the robustness section as I report selfbleu.}
% \begin{table}[t]
%     \centering
%     \begin{tabular}{l@{\;}r@{\;}r@{\;}r@{\;}r}
%     \toprule
    
%     Model&val loss&val BLEU&test loss&test BLEU\\ \midrule
%          Trans&$0.01401$&$98.079$&$7.401$&$98.078$ \\
%          NVIB &$0.06929$&$96.741$&$8.587$&$96.862$ \\
%     \bottomrule
%     \end{tabular}
%     \caption{Models performance on the reconstruction task.}
%     \label{tab:reconstruction}
% \end{table}

\subsection{Attention Map Visualisations and Analysis} \label{sec:attention}

% Figure \ref{fig:attention_sent2_1lyr} compares the self-attention patterns of the the last layer of: a Transformer with 4 layers of standard attention (left); and a Transformer with 3 layers of standard attention and 1 layer of denoising attention with NVIB (right).

\begin{figure}[t]
% layer 5 (highest layer with NVIB)
\begin{subfigure}{.237\textwidth}
  \includegraphics[width=\textwidth]{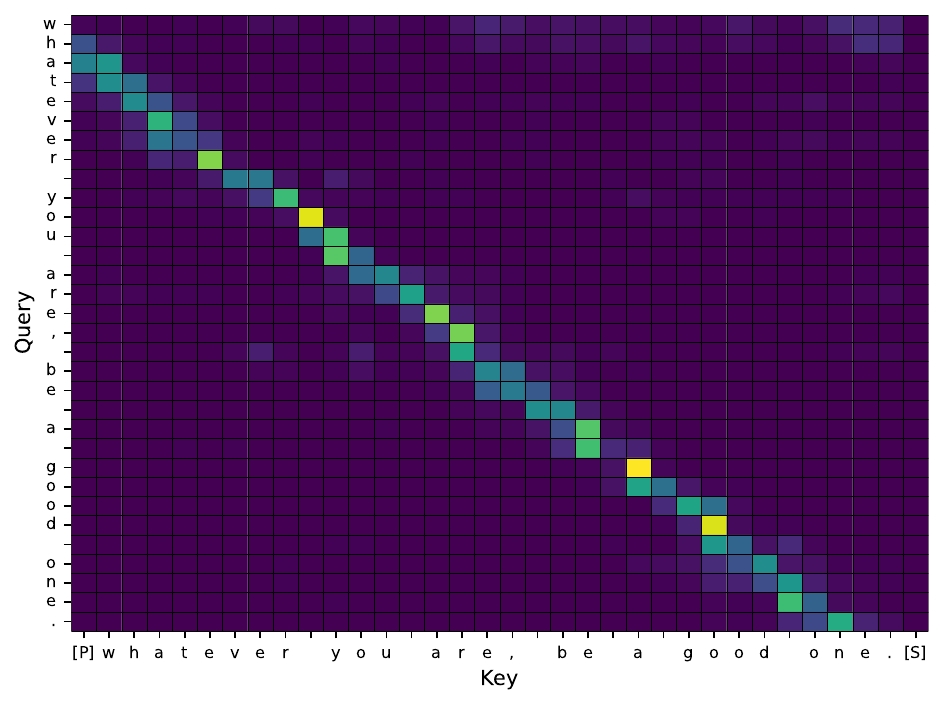}
  % \caption{}
  % \label{}
\end{subfigure} \hfill
\begin{subfigure}{.237\textwidth}
  \includegraphics[width=\textwidth]{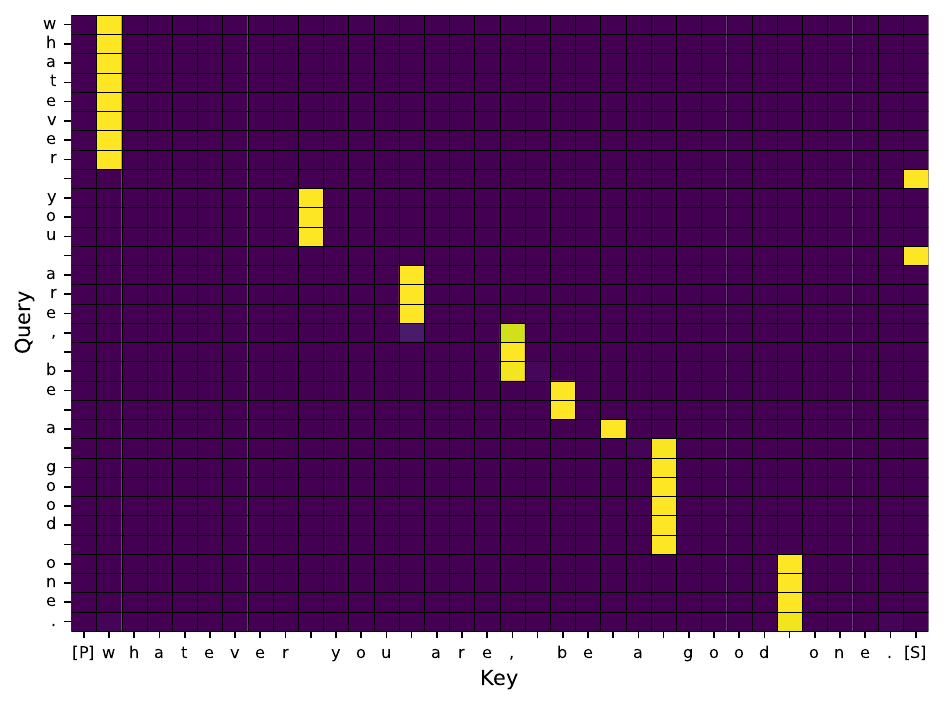}
  % \caption{}
  % \label{}
\end{subfigure}
% layer 4 (middle of 3 layers with NVIB)
\begin{subfigure}{.237\textwidth}
  \includegraphics[width=\textwidth]{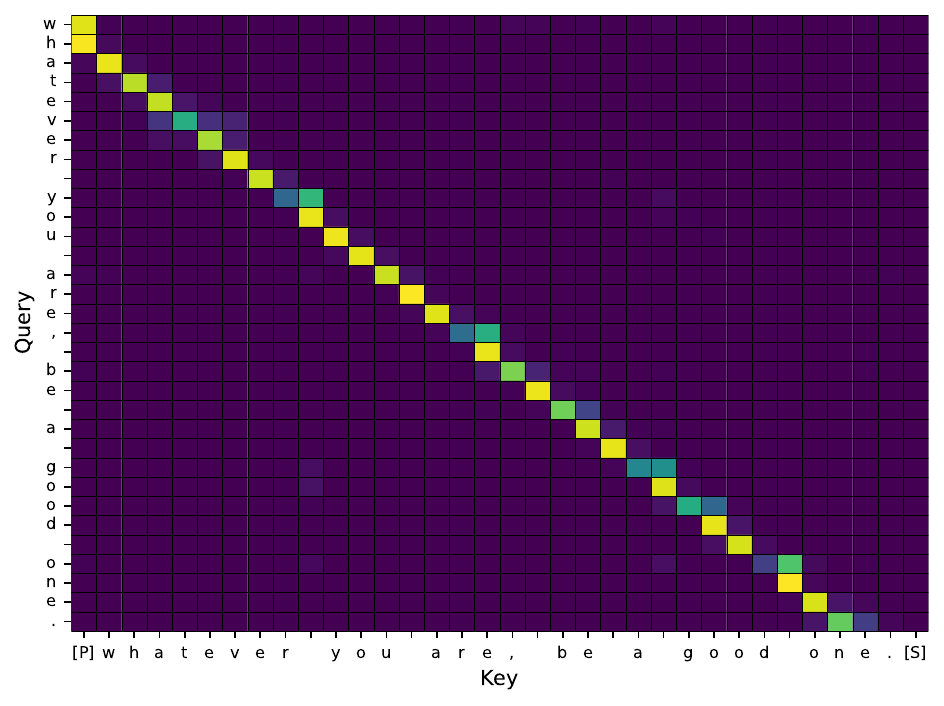}
  % \caption{}
  % \label{}
\end{subfigure} \hfill
\begin{subfigure}{.237\textwidth}
  \includegraphics[width=\textwidth]{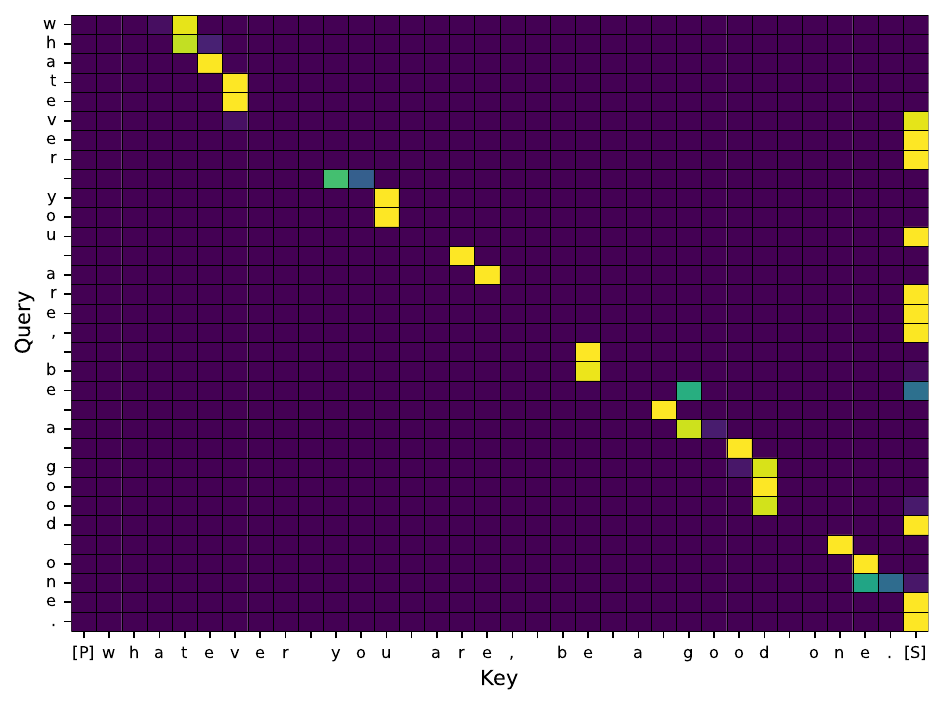}
  % \caption{}
  % \label{}
\end{subfigure}
% layer 3 (lowest layer with NVIB)
\begin{subfigure}{.237\textwidth}
  \includegraphics[width=\textwidth]{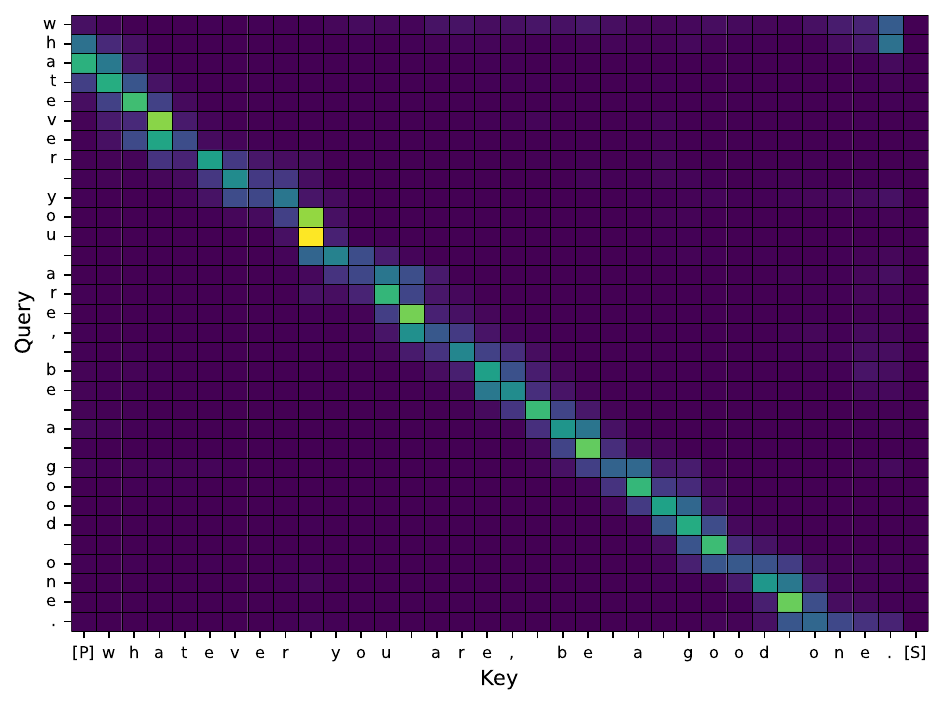}
  % \caption{}
  % \label{}
\end{subfigure} \hfill
\begin{subfigure}{.237\textwidth}
  \includegraphics[width=\textwidth]{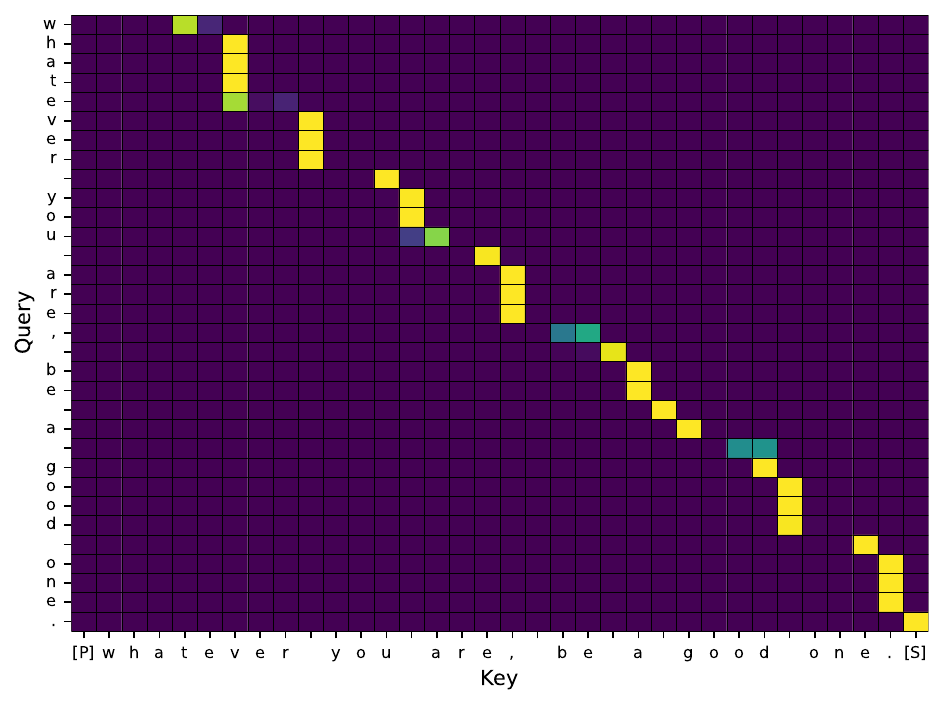}
  % \caption{}
  % \label{}
\end{subfigure}
%% layer 4 (No NVIB in this layer)
%\begin{subfigure}{.237\textwidth}
%  \includegraphics[width=\textwidth]{figures/attention_plots/T6/Attention_6_layer_2.pdf}
%  % \caption{}
%  % \label{}
%\end{subfigure} \hfill
%\begin{subfigure}{.237\textwidth}
%  \includegraphics[width=\textwidth]{figures/attention_plots/T3nvib3/Attention_6_layer_2.pdf}
%  % \caption{}
%  % \label{}
%\end{subfigure}
\caption{%Our models have 6 Transformer encoder layers; 
Self-attention patterns of the last 3 layers of 6-layer Transformer encoders from bottom to top. \textbf{Left}: Standard self-attention. \textbf{Right}: With NVIB regularisation. \textbf{Sentence}: "Whatever you are, be a good one." Dark purple is 0 and light yellow is 1 for attention.}
\label{fig:attention_sent6_3lyr}
\end{figure}

To qualitatively evaluate the model's ability to learn interpretable abstractions, we visualise the self-attention maps. Figure \ref{fig:attention_sent6_3lyr} compares the self-attention patterns of the the last %four 
3 layers of: a Transformer with 6 layers of standard attention (left); and a Transformer with 3 layers of standard attention followed by 3 layer of denoising attention with NVIB (right).

% \begin{figure}[t]
% \begin{subfigure}{.237\textwidth}
%   \includegraphics[width=\textwidth]{figures/attention_plots/T4/Attention_2_layer_3.pdf}
%   % \caption{}
%   % \label{}
% \end{subfigure} \hfill
% \begin{subfigure}{.237\textwidth}
%   \includegraphics[width=\textwidth]{figures/attention_plots/T3nvib1/Attention_2_layer_3.pdf}
%   % \caption{}
%   % \label{}
% \end{subfigure}
% \caption{\textbf{Left}: Transformer self-attention and \textbf{Right}: Transformer with NVIB with Denoising self-attention in the final layer. \textbf{Sentence}: "Wow, it's abstracting."}
% \label{fig:attention_sent2_1lyr}
% \end{figure}
% \fabio{Maybe remove this}

% Figure 2 caption that dark purple is 0 and light Yellow is 1 for the attention maps. 

Despite being trained solely on noisy reconstruction at the character level, the NVIB layers compress the self-attention representations through the layers into distinct groups. At lower levels, the model uses nearly all vectors (i.e.\ $\sim\!99\%$) and learns position-local information, shown as a diagonal pattern. At higher levels the model drops some vectors (the blank columns) and groups characters (the vertical bars) in ways which strongly resemble subword units or even words. The last level retains only an average  of $\sim\!35\%$ of vectors.   
This is because the stronger NVIB regularisation at higher layers encourages the grouping of correlated characters, to reduce redundant information, and the strongest correlations are within words. We provide further examples in Appendix \ref{sec:extraVisualisations}.
%\melika{more explanation on why the nvib model has this behavior}
%If we further increase the regularisation, the groupings become larger until all units are compressed into a single vector, much like a sentence embedding. 
%% However, a single vector is challenging to get competitive reconstruction cross-entropy during validation. 
%If we over-regularise, the representations collapse to the uninformative prior representation from NVIB.

\begin{table}[t]
    \centering
    \small
    \begin{tabular}{lrrr}
    \toprule
         &P&R&F1 \\ \midrule
         Transformer&$\textbf{95.51}$& $56.51$&$64.52$  \\
         NVIB&$85.23$& $\textbf{79.02}$&$\textbf{78.86}$\\
    \bottomrule
    \end{tabular}
    \caption{Word segmentation performance [\%].}
    \label{tab:word_seg}
\end{table}

We quantify the resemblance of the final-layer self-attention maps to words by extracting contiguous segments from the maps and computing the F1 measure between our segments and the words in the sequence. In particular, we find the best alignment between words and segments and compute the number of characters in the longest common substring between a word and its corresponding discovered segment.\footnote{See \Cref{apx:f1} for further details and exact formulas.} 
\Cref{tab:word_seg} compares the performance of our model to the Transformer baseline. This impressive unsupervised performance (F1 of $78.86\%$) concurs with the attention visualisations and quantitatively verifies that our model has learned to abstract to the level of words.
%within its layers. 

\subsection{Probing Analysis}
\label{sec:probing}

This section uses different probing tasks to quantitatively evaluate the abstraction capabilities of our model and analyse the linguistic information captured by the layers. 
%To evaluate the level of abstraction of our representations, we train probing classifiers to identify the sub-topic from short, technical input sentences \cite{hofmann-etal-2022-embarrassingly}. To analyse the linguistic features captured by the models layers, we train probing classifiers at each layer of the model and evaluate our representations on the suite of linguistic evaluations SentEval \cite{conneau-kiela-2018-senteval}. 
% For details on the probing classifiers see Appendix \ref{apx: probingclassifiers}.

% x axis layers and y is score. y is score
%  Just top const and coordINV in paper. 
% Put rest in table in appendix.

% Attention based prob idea - 
% Top const (syntax) and cooordInv (semantics) - Make argument about why words better than chars for this task. - Test against word level... noooooo

% Sentence len - word content =chars are for surface levels and not for chars. Count spaces? 

\subsubsection{ArXiv Topic Classification} \label{sec:classification}

% Rabeehs paper used VIB in low resource settings 
%  for a couple of classification tasks - sentiment classification (YELP + IMDB) and NLI (SNLI and MNLI) 

%  Then out of domain generalisation many NLI datasets

%  Hypermixer uses SST2, MNLI SNLI QQP and QNLI from GLUE.

%  Then debuggers work nicely for adverserial attacks here. 

%  Classification tasks Arxiv sub category (could fine tune or just freeze the model.)

%  Training a head on the layers for downstream tasks like sentiment analysis.

%  Less data argument. With less data we should be able to learn these properties.

% Assumption: NVIB model learns higher abstractions of the input than a normal Transformer. Thus, it should perform better on real downstream. 
The ArXiv topic classification task \cite{hofmann-etal-2022-embarrassingly} is a challenging task consisting of short input sentences with long technical words. For each subject, the classifier should classify the topic into 20 possible sub-areas. Following \citet{behjati2023inducing}, we train an attention-based probe on the final layer of the models and report the F1 measure for performance on the ArXiv-L dataset. Without finetuning the models, this classification task serves as probing high-level abstract linguistic properties \cite{hewitt-etal-2021-conditional}. 
%We provide more details on the setup in Appendix \ref{apx:arxiv}. 
%Table \ref{tab:arxiv} shows the performance of our model and the baseline.
%%\footnote{Note that we do not fine-tune our model and use this as a highly semantic probing task.}
\begin{table}[t]
    \small
    \centering
    \begin{tabular}{lcc}
    \toprule
         Task & Transformer & NVIB \\ \midrule
         Computer science & $42.33$ & $44.47$ \\
         Mathematics & $44.02$ & $47.13$ \\
         Physics & $48.83$ & $52.32$ \\
         \midrule
         \textbf{Average} & $45.06$ & $\textbf{47.97}$ \\ 
    \bottomrule
    \end{tabular}
    \caption{F1 score $[\%]$ on Arxiv-L classification task.}
    \label{tab:arxiv}
\end{table}
%
%\noindent
As shown in Table \ref{tab:arxiv}, the NVIB layer results in the model learning more information about the meaning and semantics in the abstract representations than characters and therefore provides better units for performing the task. 

\subsubsection{Linguistic Probing}
% \fabio{we report a subset of 7 sentlen, oddman and word content is too hard at charlevel}
The SentEval task set is specifically designed to examine the linguistic information available in a sentence representation at different levels, ranging from surface-level to semantic-level tasks \cite{conneau-etal-2018-cram, conneau-kiela-2018-senteval}. We probe for linguistic information of our model and the baseline Transformer, across all layers. In general, the performance improves in deeper layers and increases further with the inclusion of NVIB in the layers. 
% In general, both of the models performance improves in deeper layers. However, our proposed model works better in nearly all the reported tasks at each layer. 
% Highlighting 
%Bigram shift (\textbf{BShift)}, Coordination Inversion (\textbf{CoordInv}), Tense (\textbf{Tense}) and Top Constituents (\textbf{TopConst}) for the last 3 layers of our model in Figure \ref{fig:senteval}.
% 
\begin{figure}[t]
    \centering
    \includegraphics[width=0.65\linewidth]{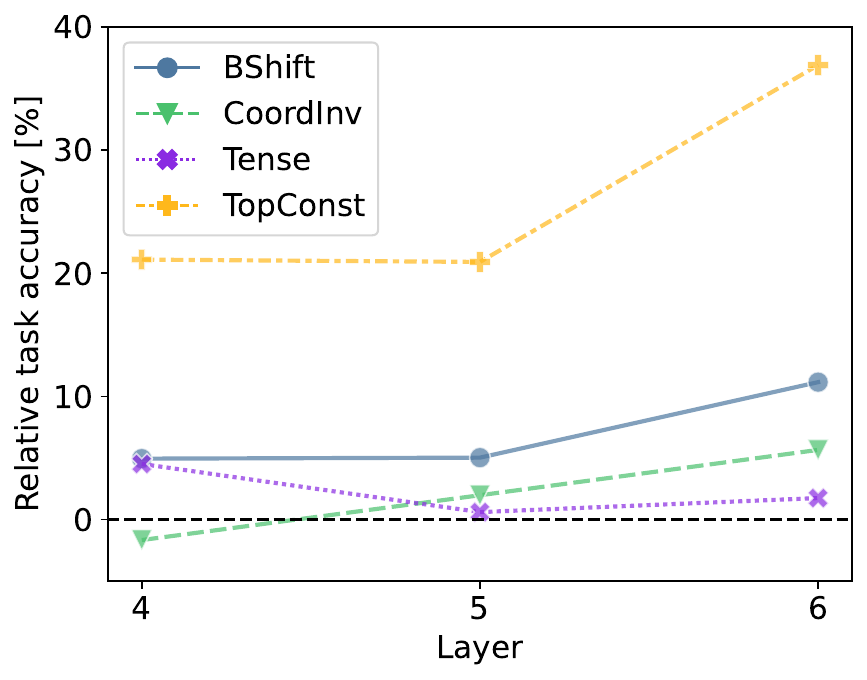}
    \caption{Relative performance of NVIB over Transformer for a subset of SentEval tasks.}
    \label{fig:senteval}
\end{figure}
%
%\noindent
We highlight the results of four tasks in Figure \ref{fig:senteval}, which to perform well in these tasks the representations must capture latent syntactic structures (\textbf{BShift}), cluster them by constituent types (\textbf{TopConst}), or have an understanding of semantics (\textbf{Tense}) or broad discourse and pragmatic factors (\textbf{CoordInv}) \cite{conneau-etal-2018-cram}. The inclusion of our NVIB layers increases the relative performance over the Transformer baseline, showing it to be more linguistically informed.  
%We report 
The complete set of results is in Appendix \Cref{tab:senteval}.

\subsection{Robustness Analysis}
\label{sec:robustness}
\begin{figure}[t]
% layer 0
\begin{subfigure}{.237\textwidth}
  \includegraphics[width=\textwidth]{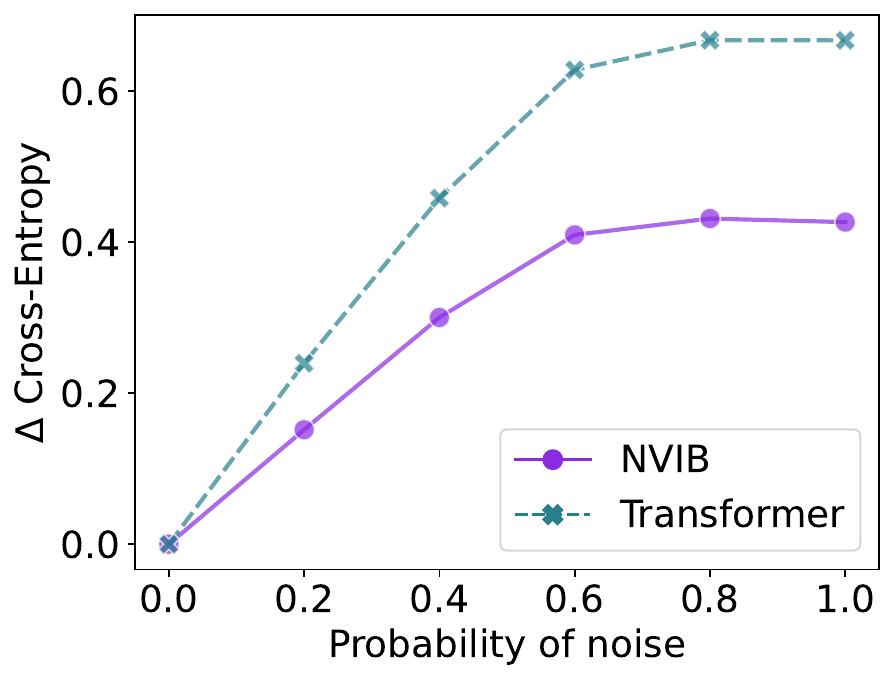}
  % \caption{Relative change in cross-entropy.}
  % \label{}
\end{subfigure} \hfill
\begin{subfigure}{.237\textwidth}
  \includegraphics[width=\textwidth]{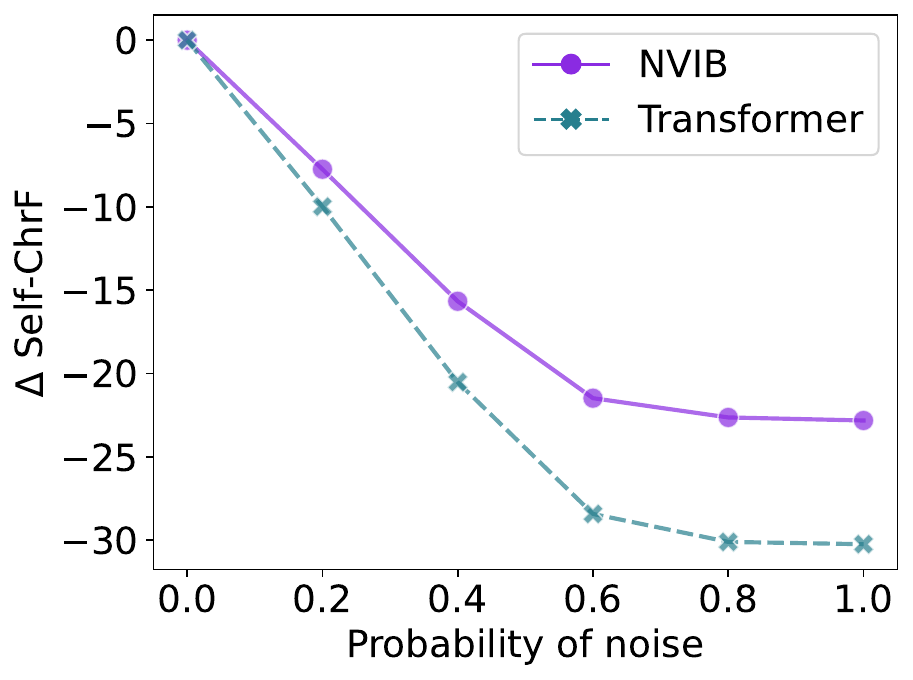}
  % \caption{Relative decrease in self-CHRF score.}
\end{subfigure}
\caption{Robustness plots showing relative performance change over increasing input perturbations.}
\label{fig:robustness}
% \caption{}
\end{figure}

We analyze the robustness of our models to synthetic noise injected into the input sequences
\cite{belinkov2017synthetic,durrani2019one}. Namely, we evaluate the reconstruction quality when the inputs are perturbed by swapping, deleting, inserting, and substituting characters \cite{morris2020textattack}.  We expect our model to be more robust due to its compressed representations. Figure \ref{fig:robustness} shows that our model is more robust to adversarial noise than a standard Transformer, with increased advantage as the level of noise increases. 
%Indeed, the final layer of our model retains only an average of 35\% of its vectors. 

% For this purpose, we use TextAttack \cite{morris2020textattack} library to craft adversarial examples by injecting synthetic noise into the original input sequence, by swapping, deleting, inserting, and substituting characters. We then compare the reconstruction quality of our proposed model in comparison to a normal Transformer on the Validation set of the training data. 
% \melika{We can put an example for this in the Appendix }
% \fabio{I think we can afford a plot here.}
% \paragraph{Results.} \Cref{fig:robustness} shows that our model is more robust to adversarial noise than a Transformer. We believe this is due to its compression abilities via the NVIB layer. 
%Building a model that is more robust to noise by design is a valuable step towards

% \begin{itemize}
%     \item Try synthetic noise as in \cite{durrani2019one,belinkov2017synthetic}, e.g., swapping or randomizing character orders \fabio{I have tried this and NVIB and Transformer are the same on character level deletion and performs worse on word level deletion. The training appears to learn to copy and not to fill in the missing tokens.}
%     \item Try textbugger \cite{li2018textbugger}
%     \item Try more advanced attacks (semantic preserving by replacing important words)
%     \item Cross entropy
% \end{itemize}

\section{Conclusions} % and Discussion}

% In this work w
We propose a novel method for inducing abstract representations of text.
We adapt the Nonparametric Variational Information Bottleneck \cite{hendersonfehr2023} regulariser for application to self-attention in the stacked layers of a Transformer encoder.  Our model learns how many vectors are needed at each layer, thereby inducing different levels of abstraction in different layers of the same model.
We find that these abstract units are intuitive, more robust, and better at encoding semantically and linguistically meaningful information.

%\begin{itemize}
%    \item novel adaptation of NVIB to self-attention layers
%    \item show its able to learn more abstract units than a normal transformer and has better representations, more robust to noise ... 
%\end{itemize}
%\melika{I think we need to discuss here what the model could be useful for. }

\section*{Limitations}

While the models and training data are reasonable in size, the experiments do not include the very large scale training often found in work on representation learning in text.  We anticipate that the advantages of NVIB on self-attention layers will only increase as the models and data are scaled up, since this should allow even more abstract representations to be learned. In addition, the experiments are only done on English, but we would expect more improvements with more morphologically rich languages. In future work we plan to explore fine-tuning NVIB for sparsity and downstream performance, and consider different tokenizations beyond characters only.  
% characters only, single language, small alphabet size, specific task-tuned models

% The current experiments are done with only one attention head per layer, whereas multi-head attention is standard.  This was done to simplify the analysis and the implementation.  A multi-head version of this model is planned for future work.

%\begin{itemize}
%    \item Large-scale training? and Pretraining? Fine-tuning?
%    \item Experiments only on English? as morphologically poor language
%    \item Future work - other tokenizers + compare against bytepair etc. (JH: doesn't learn; same at all layers)
%    \item Future work - comparison with charformer and canine.  (JH: doesn't learn)
%    \item Future work - multihead application?
%\end{itemize}

\section*{Ethics Statement}
We foresee no ethical concerns with our work. 
\section*{Acknowledgements}

Both Melika Behjati and Fabio Fehr were supported by the Swiss National Centre of Competence in Research (NCCR) under the project Evolving Language, grant number ``51NF40\_180888''. 
\bibliography{anthology,custom}
\bibliographystyle{acl_natbib}

\appendix

\section{Training Details} \label{apx:training}

\paragraph{General Training} All models are trained, without pretraining, using the same encoder and decoder configuration for comparability. Our encoder size is defined by the base Transformer \cite{vaswani2017} such that we have a six layer Transformer encoder. However, we use a two layer decoder to ensure the task is not learned in the decoder alone. We use a single attention head.
%as NVIB is currently only defined for this. 
The size for the word embedding vectors and model projections are $512$ and feed forward dimensions $512$, which leads to models of approximately $12$-$17$ million trainable parameters. An English character level tokeniser is used for tokenisation with a vocabulary of approximately 100 characters. During training we use: a learning rate of $1e^{-3}$ with a cosine cool-down over all steps, RAdam optimiser \cite{Liu2020On} with mixed precision (FP16), a batch size of $512$, gradient norm clipping $0.1$ and trained for $55$ epochs ($\approx8K$ steps). The number of steps were selected considering model convergence and minimising computation time. We use a dropout rate of $0.1$. The input is noised at each batch with a probability of character deletion of $0.1$. Each model takes approximately $2.5$hrs on a single NVIDIA GeForce RTX 3090.

\paragraph{NVIB Training}

Training the models with the NVIB layers requires regularising the representations. The introduction of the exponential activation function (as opposed to ReLU) for the psuedo-count parameter $\boldsymbol{\alpha}$ requires a threshold at test time to be exactly zero. We use a threshold for this at $0.1$. During training and testing we enforce a bottleneck between the encoder and decoder by masking the final encoder representations by the aforementioned threshold.

The NVIB hyperparameters $\lambda_G$, $\lambda_D$ and $\alpha^\Delta$ are selected through hyperparameter tuning. However, during training we only sample once from each component thus the approximation parameter is set to $\kappa^{\Delta}=1$. We use a Kullback-Leibler annealing divergence strategy where the introduction of the KL divergence loss is linearly introduced between $30\% - 60\%$ of the training steps. This allows the model to learn initial representations, slowly introduce the regularisation and finally learn through the compressed latent representation.

\section{Hyperparameter Tuning} \label{apx:tuning}

The models are trained on the Wikitext-2 training dataset using the loss from Equation~\ref{eq:loss}. They are tuned on the validation dataset with the aim to be able to reconstruct the character sequence from a noisy input. Following from \cite{hendersonfehr2023} we set the weights of the Gaussian and Dirichlet KL divergences to be independent of the sentence length $n$ and dimensionality of vectors $d$:
\begin{align}
\lambda_D &= \frac{1}{n} \lambda^\prime_D \ \ ; \ \  \ \ 
\lambda_G = \frac{1}{d} \frac{1}{n} \lambda^\prime_G \nonumber
\end{align}
where $\lambda^\prime_D$ and $\lambda^\prime_G$ are fixed hyperparameters. All combinations of the following hyperparameters were considered in a grid search for the respective models:
\begin{itemize}
    \item $lr = \{1e^{-4}, 1e^{-3}\}$
    \item $\lambda_G^\prime = \{1e^{-5}, \ 1e^{-4}, \ 1e^{-3}, \ 1e^{-2} \}$
    \item $\lambda_D^\prime = \{1e^{-2}, \ 1e^{-1}, \ 1 \}$
    \item $\alpha^\Delta \ \ = \{0 \ , 0.05, \ ..., \ 0.45, \ 0.5\}$
\end{itemize}
where $\lambda_G^\prime$ and $\lambda_D^\prime$ are the weights on the Gaussian and Dirichlet KL divergences. The $\alpha^\Delta$ represents the conditional prior parameter. The final models' hyperparameters are reported in Table \ref{tab:Hyperparms} where the validation cross-entropy (CE) is matched for NVIB and baseline Transformers.

\begin{table}[!ht]
    \centering
    \begin{tabular}{lcc}
    \toprule
          & Transformer & NVIB \\
          \midrule
        NVIB layers & - & 3 \\
        $\lambda_G$ & - & $1e^{-2}$ \\
        $\lambda_D$ & - & 1 \\
        $\alpha^{\Delta}$ & - & 0.25 \\
        Training Steps & $2.5K$ & $8K$ \\
         \midrule
         Val. CE  & 0.19 & 0.19 \\
         % Val. Self ChrF & & & & \\
         % Val. Self BLEU & & & & \\
    \bottomrule
    \end{tabular}
    \caption{Hyperparameters for final models evaluated.}
    \label{tab:Hyperparms}
\end{table}

The encoders 6 layers are inspired by the base model of \citet{vaswani2017}. For the Transformer decoder we use only 2 layers such that the decoder is not able to overpower the embeddings from the encoder it sees through cross attention.

\paragraph{NVIB Configuration}
For the NVIB layers during experimentation we considered: All layer including NVIB; the last 3 layers including NVIB; and only the final layer including NVIB. When all layers were included it was challenging to get both compression and performance as the regularisation was too strong. Only regularising the last layer managed to reduce the number of vectors but often converged to a single sentence vector with lower, non-comparable validation cross-entropy. Finally, we settled on only regularising the last 3 layers as it gave the model enough flexibility in the lower layers and progressive compression in the higher layers.

\section{KL Divergence Loss} \label{apx: kl_loss}

\citet{hendersonfehr2023} define the Kullback-Leibler divergence for NVIB with two terms: the $L_D$ for the Dirichlet distribution weights defined by $\boldsymbol{\alpha}$; and the $L_G$ for the distribution of vectors $\boldsymbol{Z}$ generated by the Gaussian components. We set the approximation parameter $\kappa_0=1$. This gives us the following loss terms for the KL divergence, where $\Gamma$ is the gamma function and $\psi$ is the digamma function:
\begin{dmath}
  \label{eq:KLloss-kappa}
  L_D
  = \log\Gamma(\alpha^q_0) -\log\Gamma(\alpha^{p^\prime}_0) \nonumber \\
  +(\alpha^q_0-\alpha^{p^\prime}_0) \left(
  -\psi(\alpha^q_0) +\psi({\alpha^q_0})
  \right) \nonumber \\
  +\left( \log\Gamma({\alpha^{p^\prime}_0}) -\log\Gamma({\alpha^q_0}) \right)
  \nonumber 
\end{dmath}
where, $\alpha^q_0$ is the sum of all $\boldsymbol{\alpha}$ parameters generated by the NVIB layer. The conditional prior $\alpha^{p^\prime}_0 = \alpha^p_0 + n\alpha^\Delta$ is controlled by $\alpha^p_0=1$ and extra pseudo-counts defined by the length $n$ and a hyperparameter $\alpha^\Delta$. The KL divergence between two Gaussians (with diagonal covariance with values $\boldsymbol{\sigma}$ and weighted by the $\boldsymbol{\alpha}$ parameters) is:
\begin{dmath}
  L_G = \tfrac{1}{2} \sum_{i=1}^{n+1} ~\frac{{\alpha}^q_i}{\alpha^q_0}~  \sum_{h=1}^d \left( \frac{({\mu}^q_{ih}-{\mu}^p_{h})^2}{({\sigma}^p_{h})^2} -1 \nonumber \\ + \frac{({\sigma}^q_{ih})^2}{({\sigma}^p_{h})^2}
  -\log\frac{({\sigma}^q_{ih})^2}{({\sigma}^p_{h})^2} \right)
  \nonumber
\end{dmath}

% If we know the input length $n$, but know nothing about the content of the text, then the distribution of vectors should stay the same as the general prior, $G^{p^\prime}_0=G^p_0$.  However, the count of observations we expect to have after an input of that length would not be $\alpha^p_0$, but should include a pseudo-count $\alpha^\Delta\in\mathbb{R}_{\geq 0}$ hyperparameter for every token, and thus
% $ %\[
% \alpha^{p^\prime}_0 = \alpha^p_0 + n\alpha^\Delta
% $. %\]
% This then gives us the conditional prior given $n$ of $\BDP(G^p_0,\alpha^{p^\prime}_0,\kappa_0)$.

\section{Denoising attention function} \label{apx:dattn}

\citet{hendersonfehr2023} generalise the set of vectors input to an attention function to a probability distribution over vectors, and generalise attention to a function of these probability distributions called denoising attention.

\paragraph{Training function}
During training, the set of sampled vectors $\boldsymbol{Z} \in \mathbb{R}^{n \times p}$ and their sampled log-probability weights $\text{log}(\boldsymbol{\pi}) \in \mathbb{R}^{1 \times n}$ are both output by the NVIB layer, thereby specifying the sampled mixture distribution $F$. A set of query vectors $\boldsymbol{u^\prime} \in \mathbb{R}^{m \times p}$ is projected into key space by the grouped matrices $\boldsymbol{W}^Q, \boldsymbol{W}^K \in \mathbb{R}^{p \times d}$ to $\boldsymbol{u} = (\boldsymbol{u}^\prime \boldsymbol{W}^Q (\boldsymbol{W}^K)^T)$.  The keys' dimensionality $d$ is used for scaling.  Denoising attention can then be computed as:
\begin{dmath}
 \text{softmax}\left( \tfrac{1}{\sqrt{d}}\boldsymbol{u}{\boldsymbol{Z}}^T +\log(\boldsymbol{\pi}) -\tfrac{1}{2\sqrt{d}}\|\boldsymbol{Z}\|^2 \right) \boldsymbol{Z}
\end{dmath}
For self-attention, we define the original query vectors $\boldsymbol{u^\prime}$ to be the set of vectors input to the NVIB layer, before projecting to DP parameters and sampling.
%$\boldsymbol{Z}$.

\paragraph{Testing function}
During test time, we do not sample $F$, but instead use the mean of the posterior distribution. The test-time denoising attention can then be computed as:

\begin{dmath}
\text{softmax}\left(\!\! \boldsymbol{u} \left(\frac{\boldsymbol{\mu}}{(\boldsymbol{\sigma}^r)^2}\right)^{T} +\log(\frac{\boldsymbol{\alpha}}{\alpha_0}) \\ -\left(\tfrac{1}{2}\left\|\frac{\boldsymbol{\mu}}{\boldsymbol{\sigma}^r}\right\|^2\right)^{T} -\boldsymbol{1}_p\left(\log(\boldsymbol{\sigma}^r)\right)^T \right) \\ \times \left( \frac{(\boldsymbol{\sigma})^{2}}{(\boldsymbol{\sigma}^r)^2}\odot(\boldsymbol{1}_n^T\boldsymbol{u}) +\frac{\sqrt{d}}{(\boldsymbol{\sigma}^r)^2}\odot\boldsymbol{\mu} \right)
\end{dmath}
where $\boldsymbol{1}_p$ is a row vector of $p$ ones and  $(\boldsymbol{\sigma}^r)^2=(\sqrt{d}+(\boldsymbol{\sigma}^q)^2)$.

\section{Visualisations} \label{sec:extraVisualisations}

\begin{figure}[!ht]
% layer 5 (highest layer with NVIB)
\begin{subfigure}{.237\textwidth}
  \includegraphics[width=\textwidth]{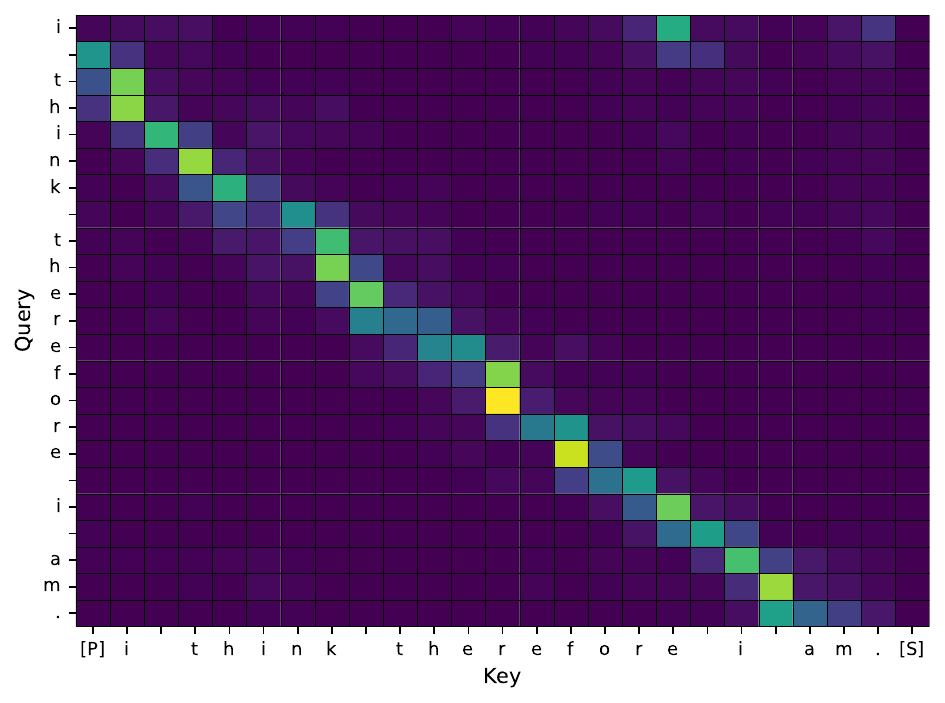}
  % \caption{}
  % \label{}
\end{subfigure} \hfill
\begin{subfigure}{.237\textwidth}
  \includegraphics[width=\textwidth]{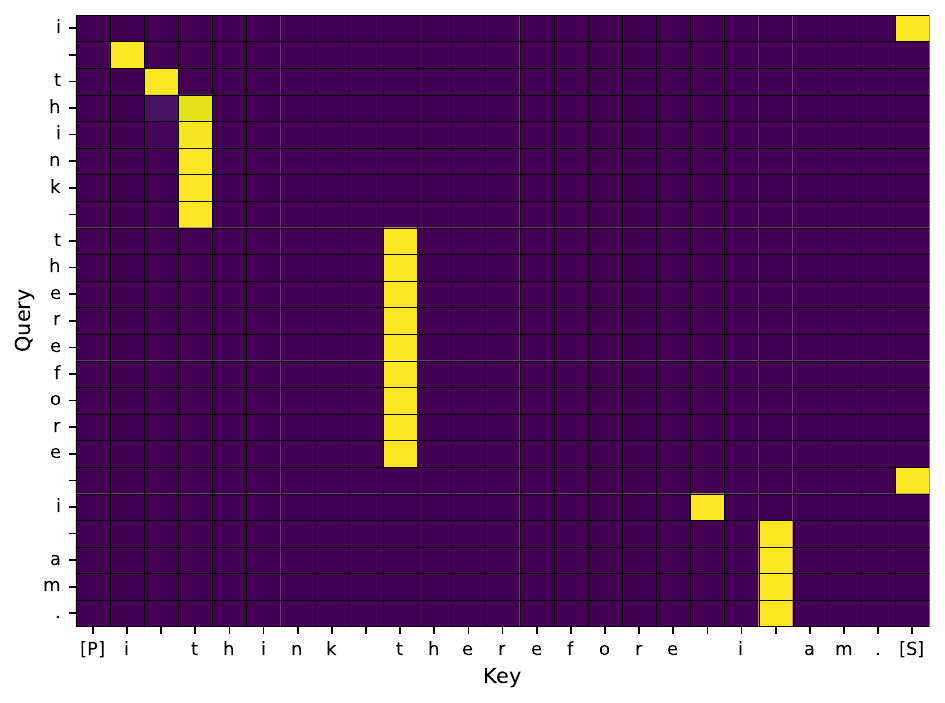}
  % \caption{}
  % \label{}
\end{subfigure}
% layer 4 (middle of 3 layers with NVIB)
\begin{subfigure}{.237\textwidth}
  \includegraphics[width=\textwidth]{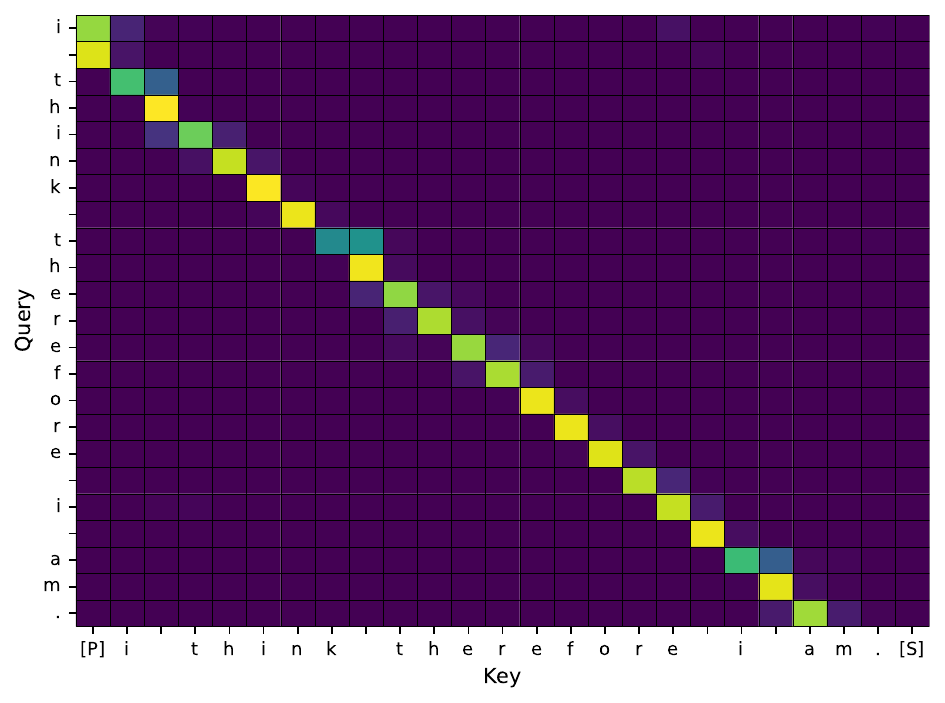}
  % \caption{}
  % \label{}
\end{subfigure} \hfill
\begin{subfigure}{.237\textwidth}
  \includegraphics[width=\textwidth]{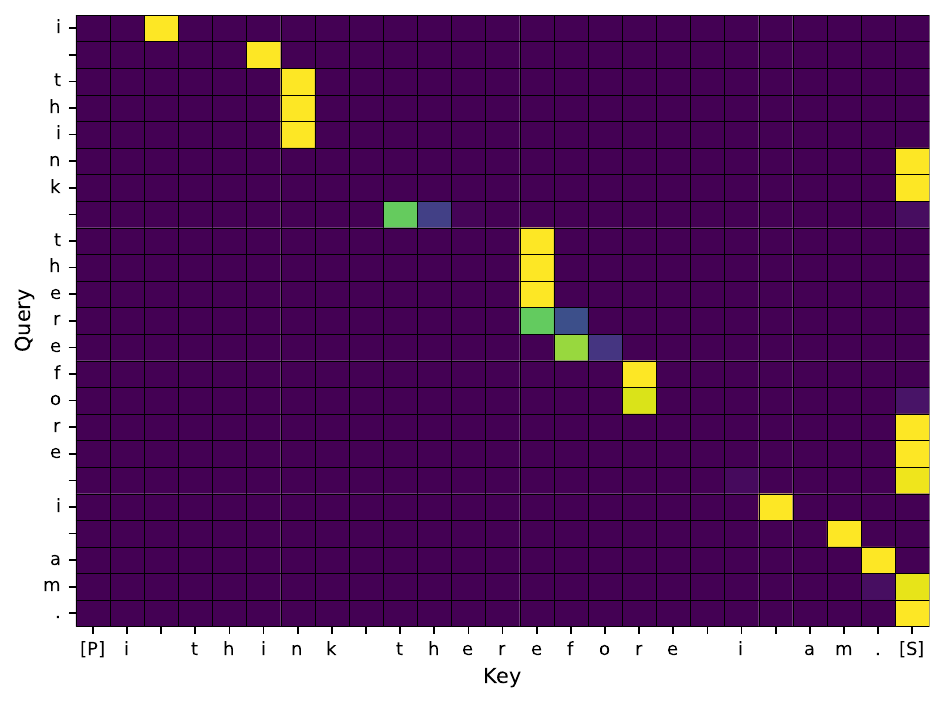}
  % \caption{}
  % \label{}
\end{subfigure}
% layer 3 (lowest layer with NVIB)
\begin{subfigure}{.237\textwidth}
  \includegraphics[width=\textwidth]{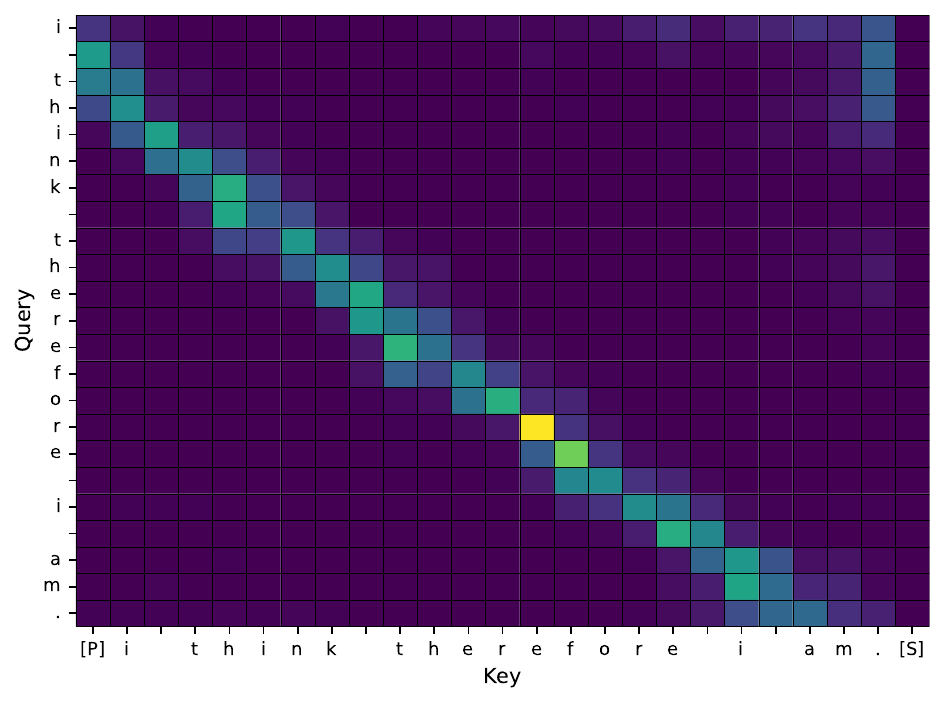}
  % \caption{}
  % \label{}
\end{subfigure} \hfill
\begin{subfigure}{.237\textwidth}
  \includegraphics[width=\textwidth]{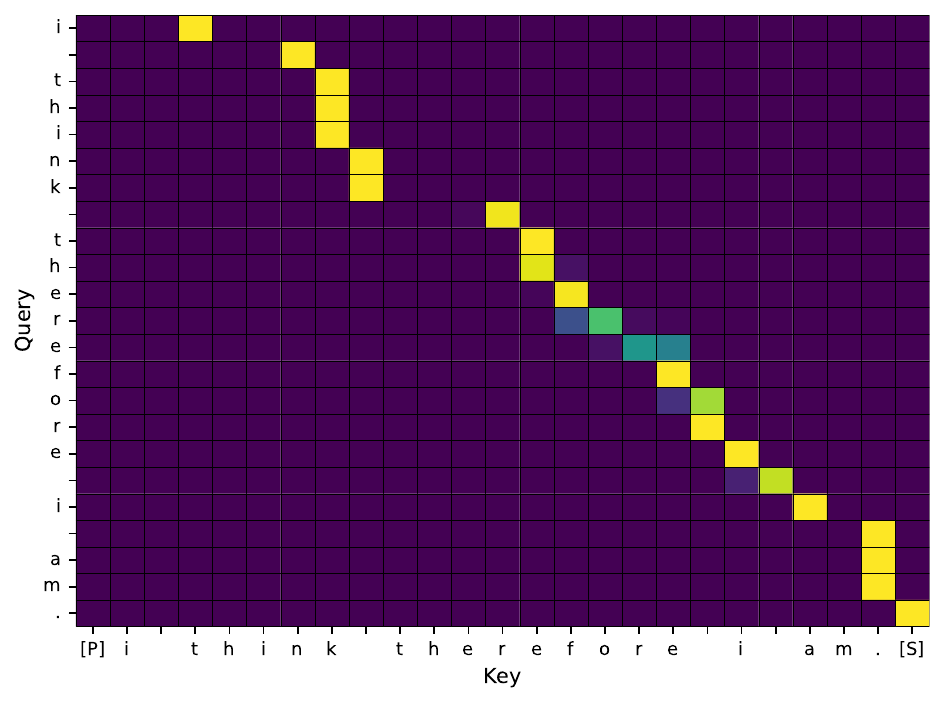}
  % \caption{}
  % \label{}
\end{subfigure}
\caption{Self-attention patterns of the last 3 layers of 6-layer Transformer encoders. \textbf{Left}: Standard self-attention. \textbf{Right}: With NVIB regularisation. \textbf{Sentence}: "I think therefore I am." Dark purple is 0 and light yellow is 1 for the attention values.}
\label{fig:attention1}
\end{figure}

\begin{figure}[!ht]
% layer 5 (highest layer with NVIB)
\begin{subfigure}{.237\textwidth}
  \includegraphics[width=\textwidth]{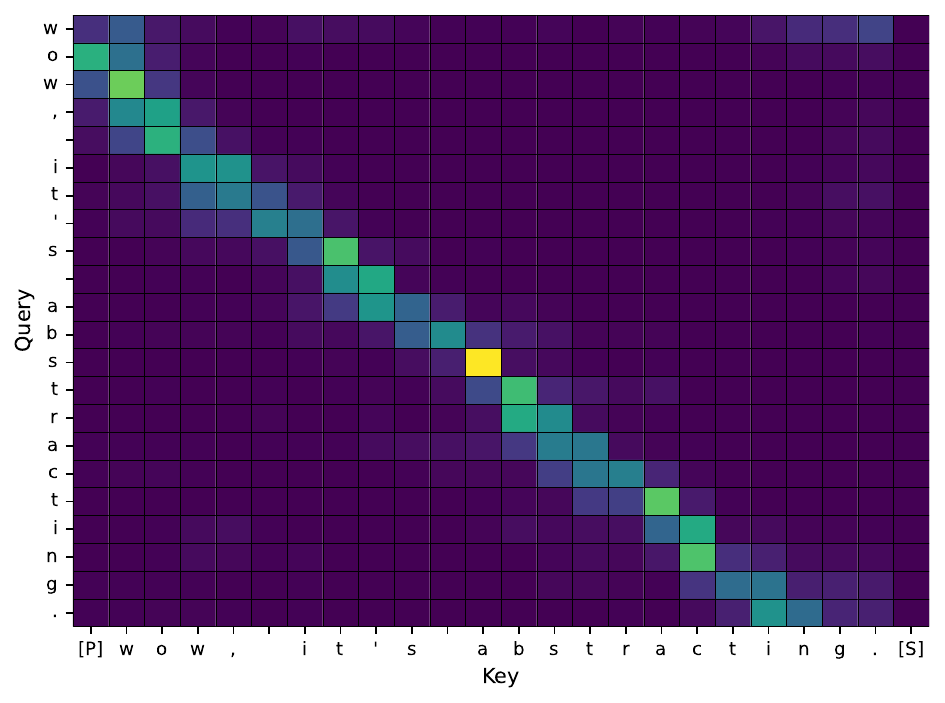}
  % \caption{}
  % \label{}
\end{subfigure} \hfill
\begin{subfigure}{.237\textwidth}
  \includegraphics[width=\textwidth]{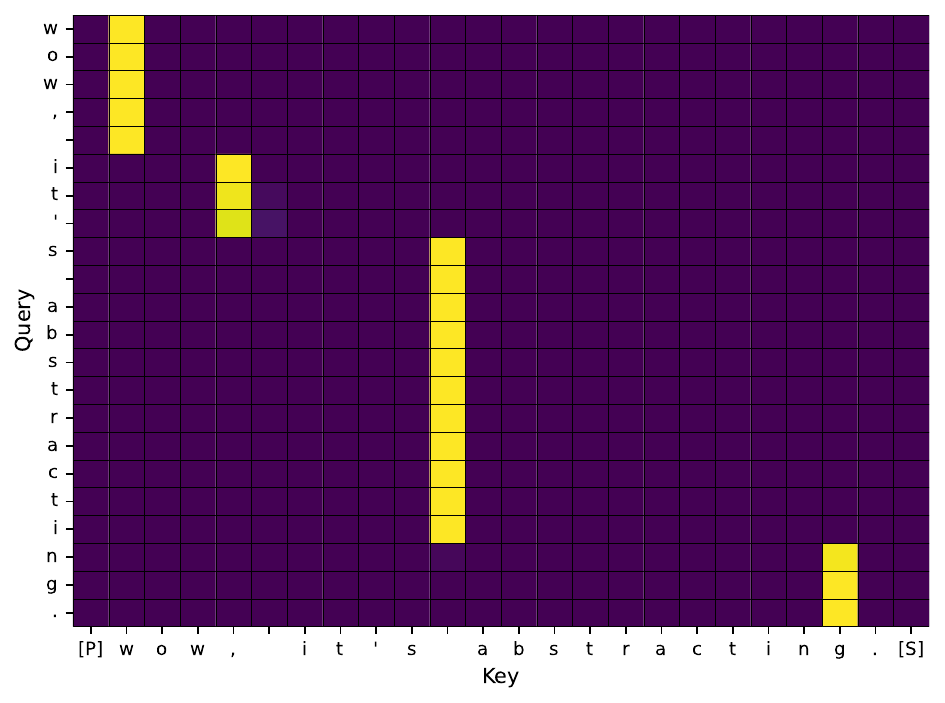}
  % \caption{}
  % \label{}
\end{subfigure}
% layer 4 (middle of 3 layers with NVIB)
\begin{subfigure}{.237\textwidth}
  \includegraphics[width=\textwidth]{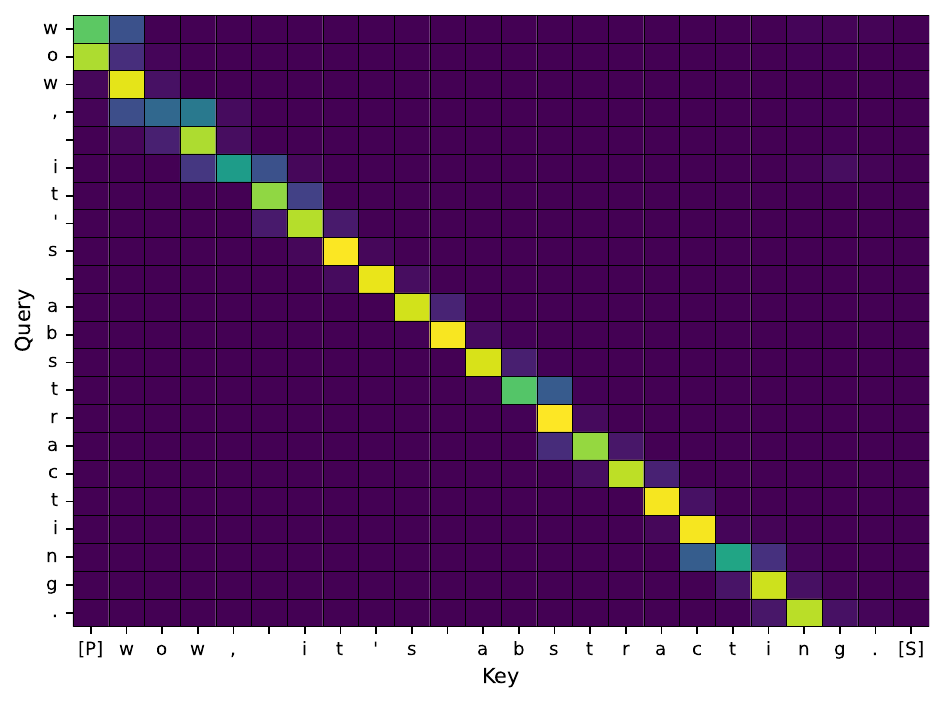}
  % \caption{}
  % \label{}
\end{subfigure} \hfill
\begin{subfigure}{.237\textwidth}
  \includegraphics[width=\textwidth]{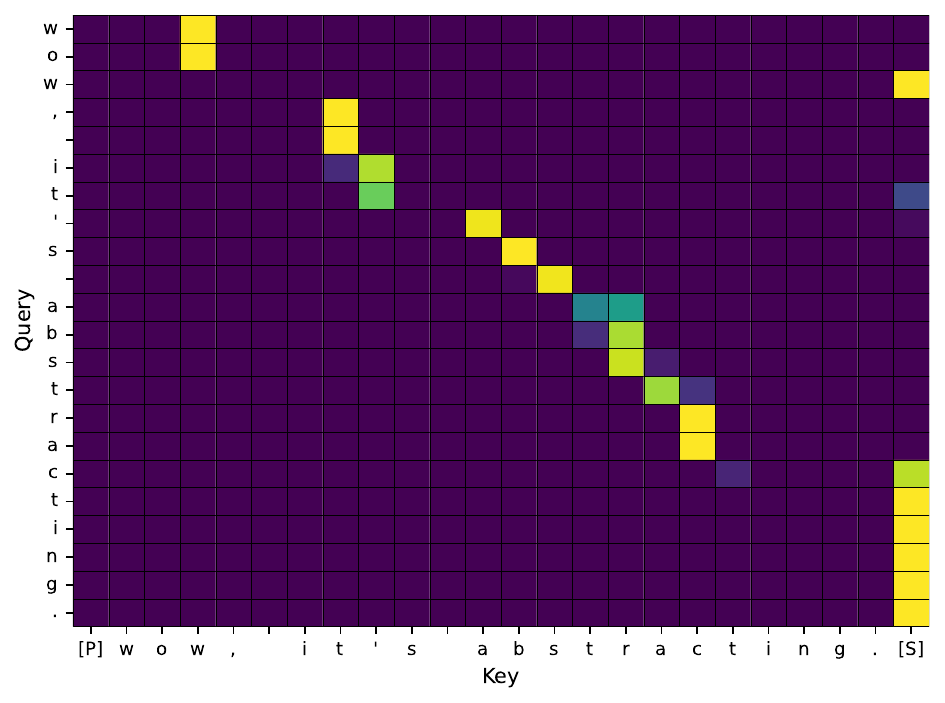}
  % \caption{}
  % \label{}
\end{subfigure}
% layer 3 (lowest layer with NVIB)
\begin{subfigure}{.237\textwidth}
  \includegraphics[width=\textwidth]{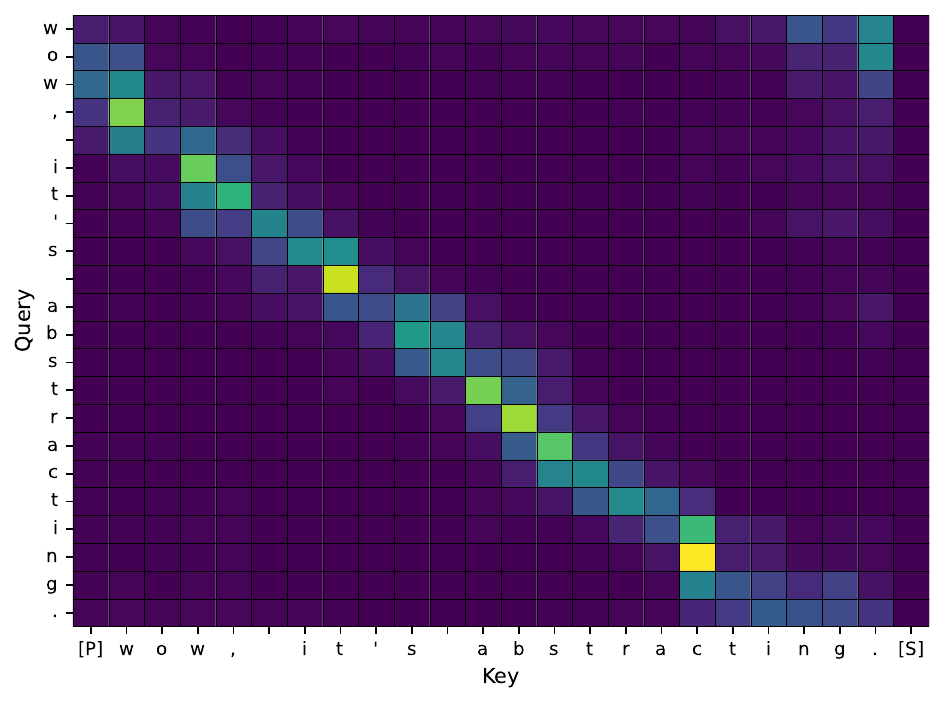}
  % \caption{}
  % \label{}
\end{subfigure} \hfill
\begin{subfigure}{.237\textwidth}
  \includegraphics[width=\textwidth]{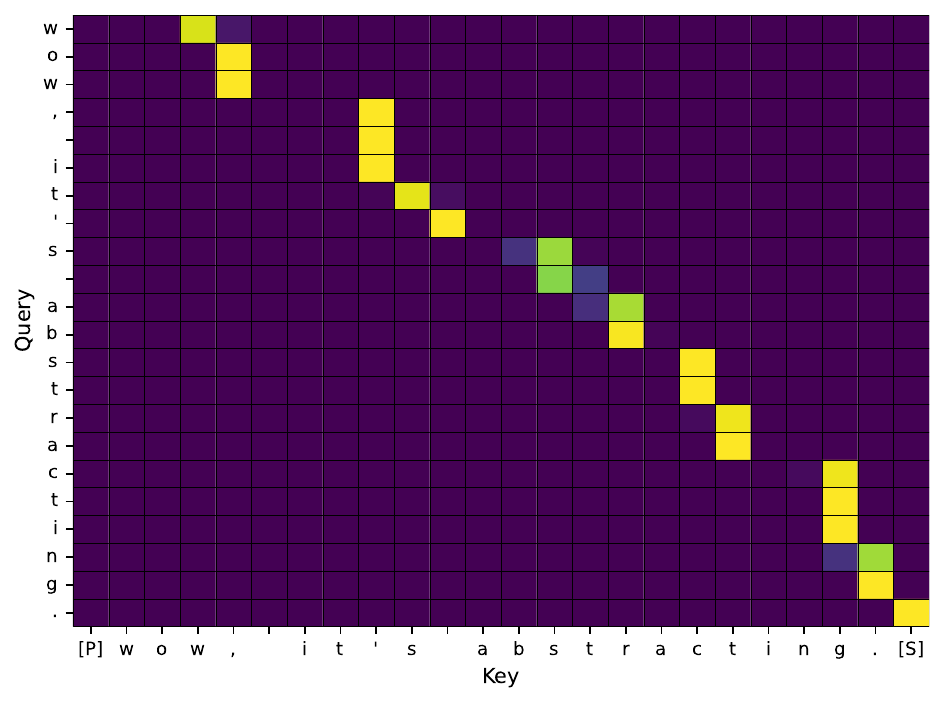}
  % \caption{}
  % \label{}
\end{subfigure}
\caption{Self-attention patterns of the last 3 layers of 6-layer Transformer encoders. \textbf{Left}: Standard self-attention. \textbf{Right}: With NVIB regularisation. \textbf{Sentence}: "Wow, it's abstracting." Dark purple is 0 and light yellow is 1 for the attention values.}
\label{fig:attention2}
\end{figure}

\begin{figure}[!ht]
% layer 5 (highest layer with NVIB)
\begin{subfigure}{.237\textwidth}
  \includegraphics[width=\textwidth]{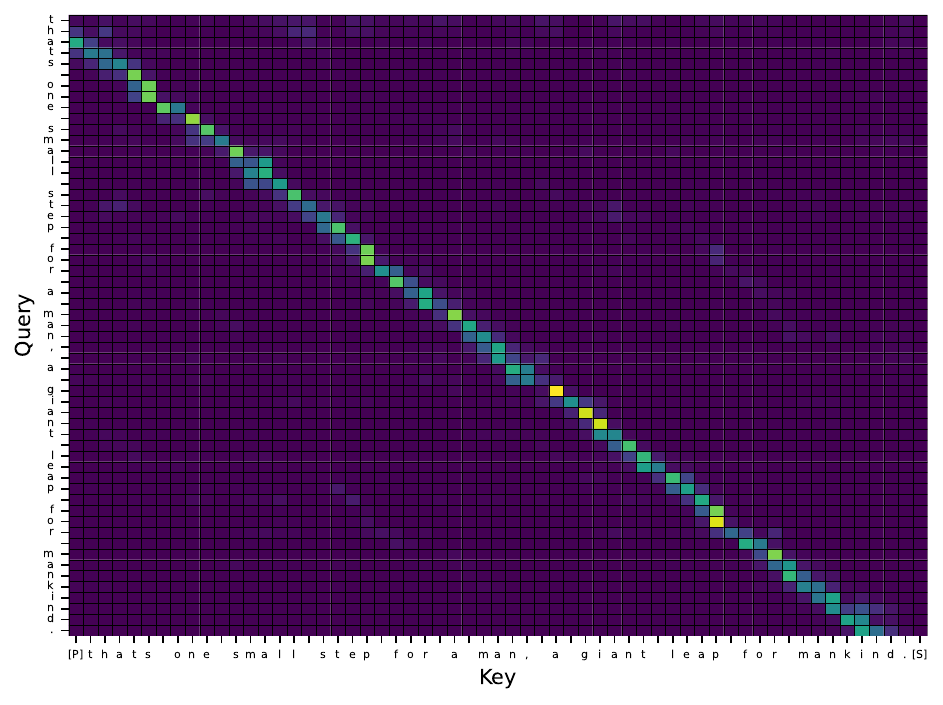}
  % \caption{}
  % \label{}
\end{subfigure} \hfill
\begin{subfigure}{.237\textwidth}
  \includegraphics[width=\textwidth]{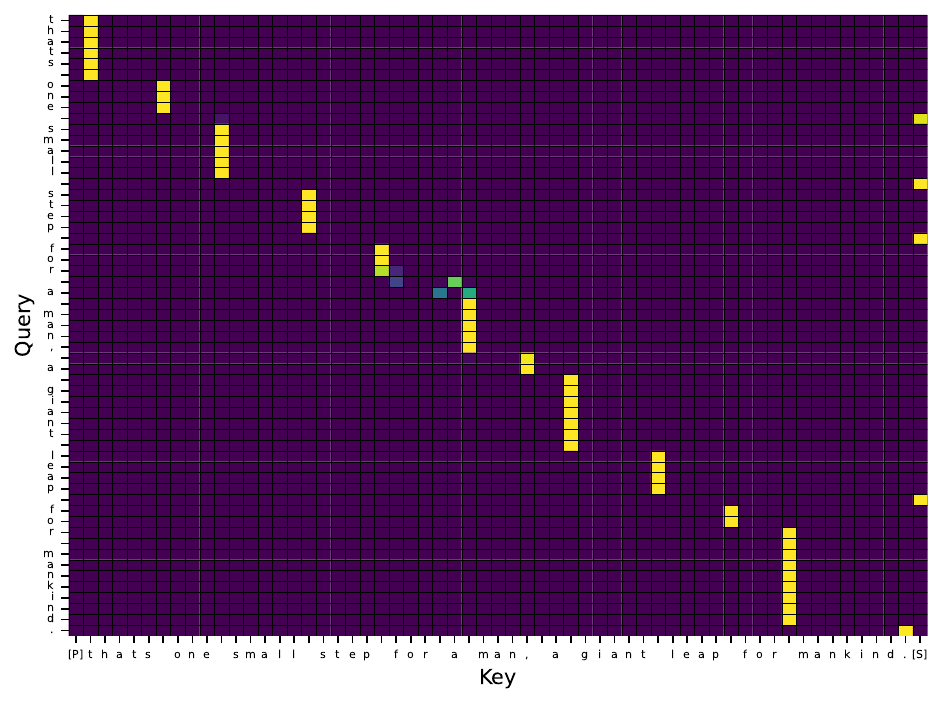}
  % \caption{}
  % \label{}
\end{subfigure}
% layer 4 (middle of 3 layers with NVIB)
\begin{subfigure}{.237\textwidth}
  \includegraphics[width=\textwidth]{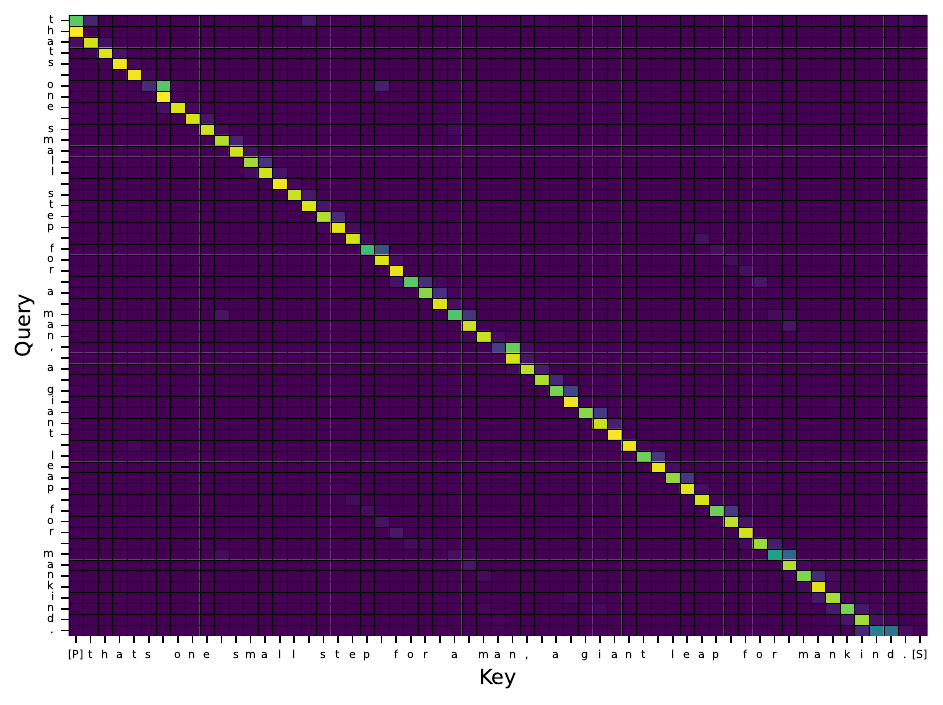}
  % \caption{}
  % \label{}
\end{subfigure} \hfill
\begin{subfigure}{.237\textwidth}
  \includegraphics[width=\textwidth]{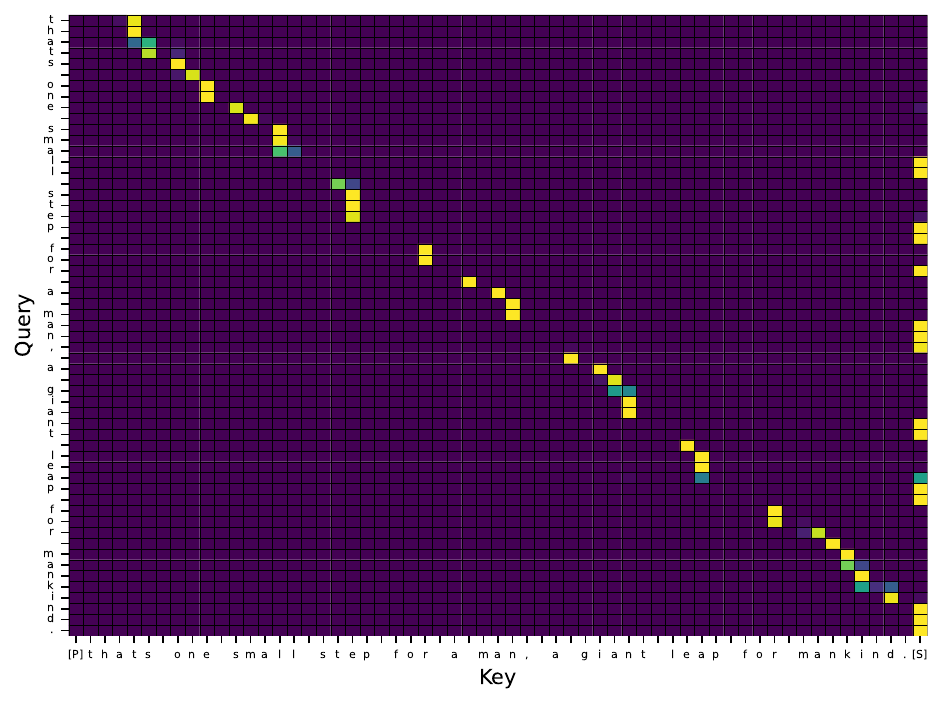}
  % \caption{}
  % \label{}
\end{subfigure}
% layer 3 (lowest layer with NVIB)
\begin{subfigure}{.237\textwidth}
  \includegraphics[width=\textwidth]{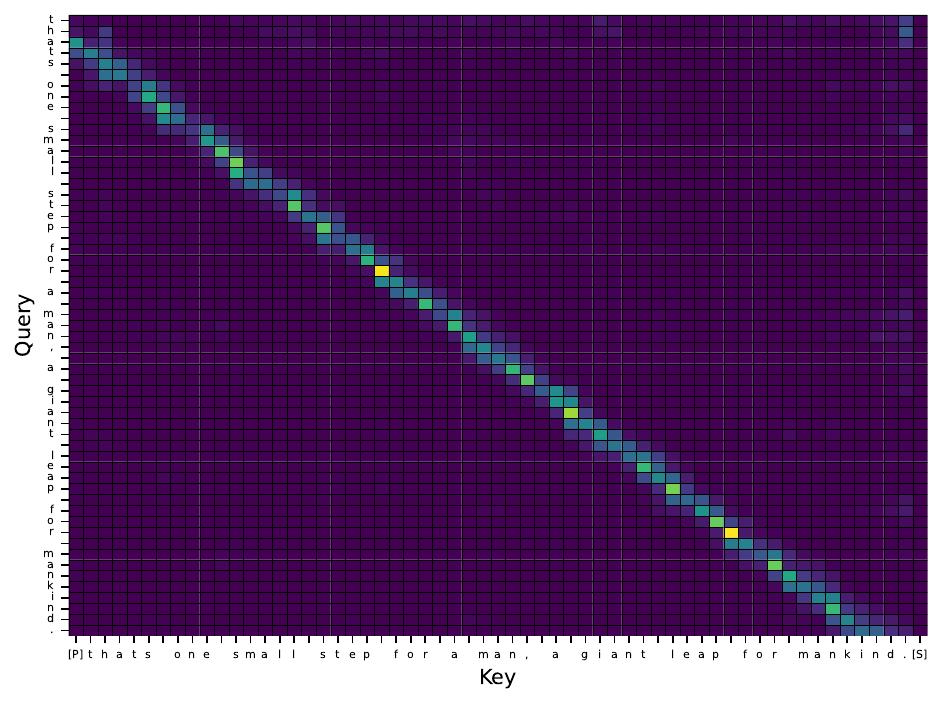}
  % \caption{}
  % \label{}
\end{subfigure} \hfill
\begin{subfigure}{.237\textwidth}
  \includegraphics[width=\textwidth]{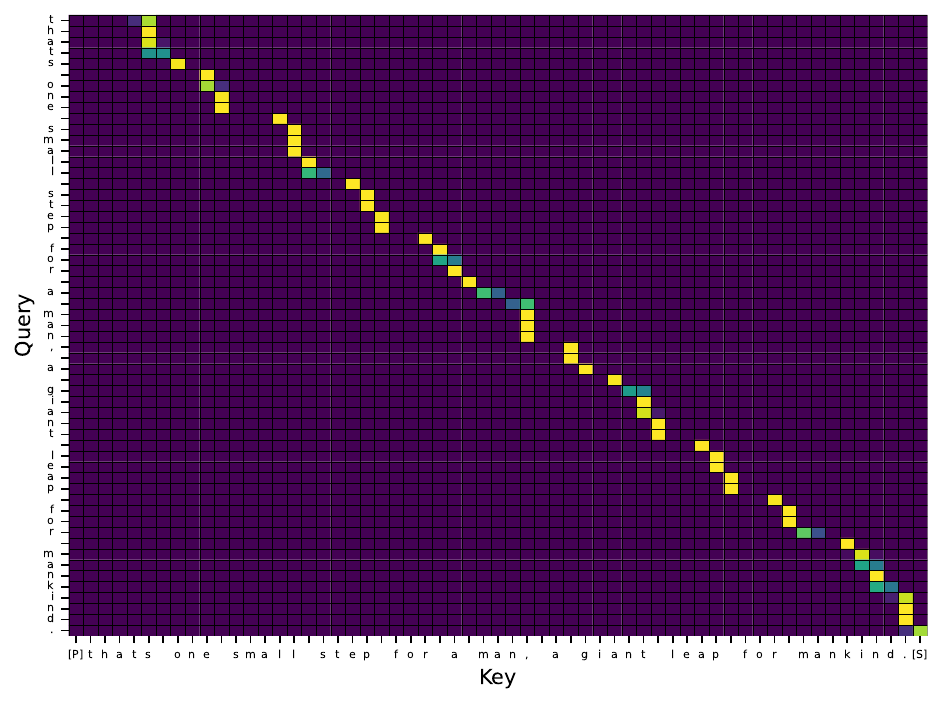}
  % \caption{}
  % \label{}
\end{subfigure}
\caption{Self-attention patterns of the last 3 layers of 6-layer Transformer encoders. \textbf{Left}: Standard self-attention. \textbf{Right}: With NVIB regularisation. \textbf{Sentence}: "Thats one small step for a man, a giant leap for mankind." Dark purple is 0 and light yellow is 1 for the attention values.}
\label{fig:attention4}
\end{figure}

\begin{figure}[!ht]
% layer 5 (highest layer with NVIB)
\begin{subfigure}{.237\textwidth}
  \includegraphics[width=\textwidth]{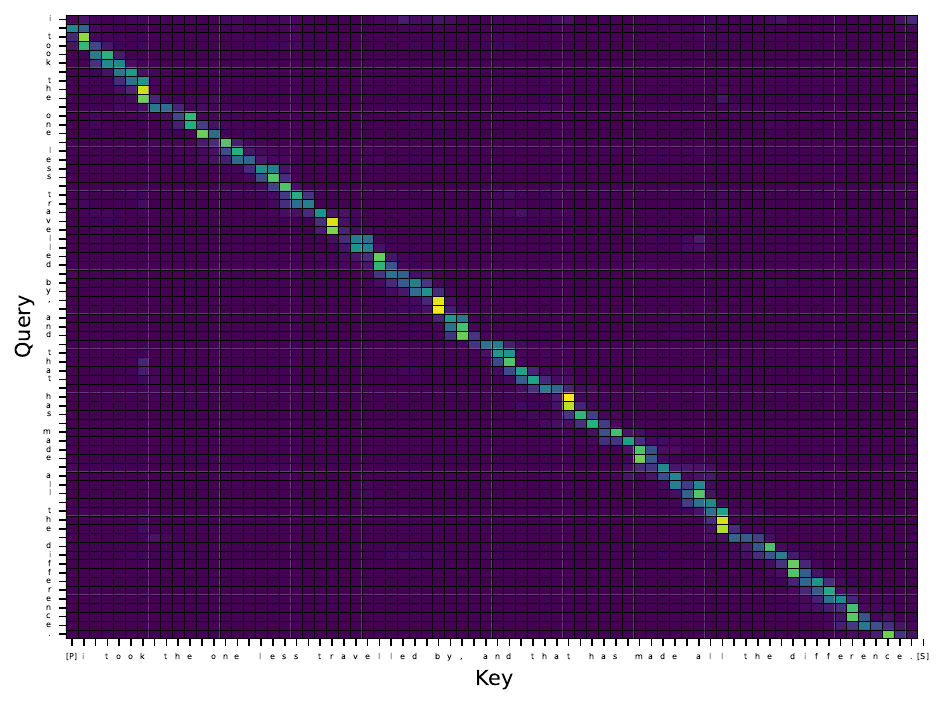}
  % \caption{}
  % \label{}
\end{subfigure} \hfill
\begin{subfigure}{.237\textwidth}
  \includegraphics[width=\textwidth]{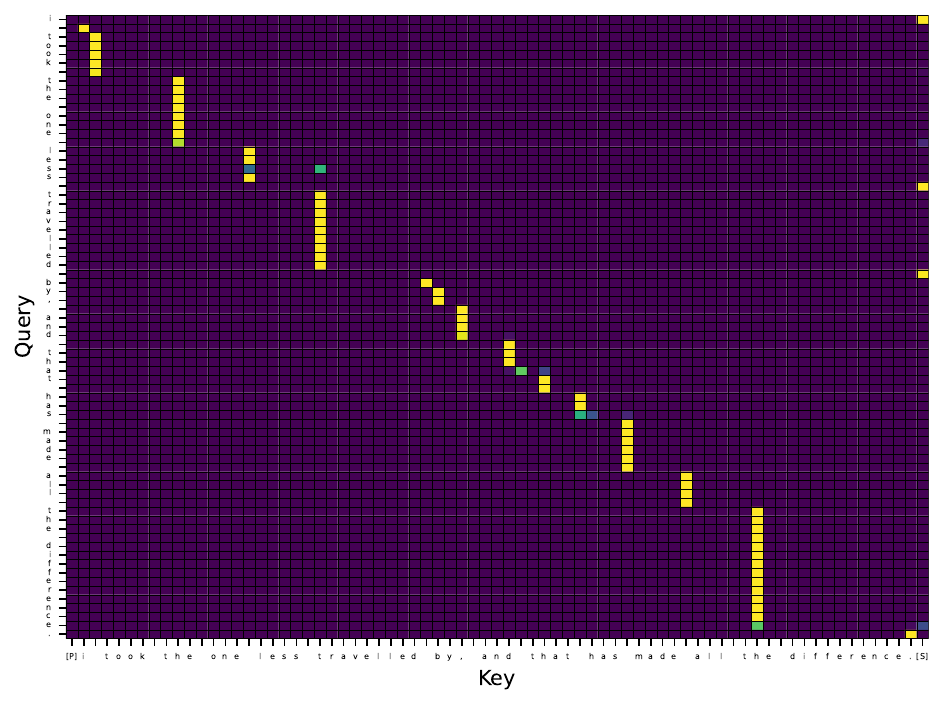}
  % \caption{}
  % \label{}
\end{subfigure}
% layer 4 (middle of 3 layers with NVIB)
\begin{subfigure}{.237\textwidth}
  \includegraphics[width=\textwidth]{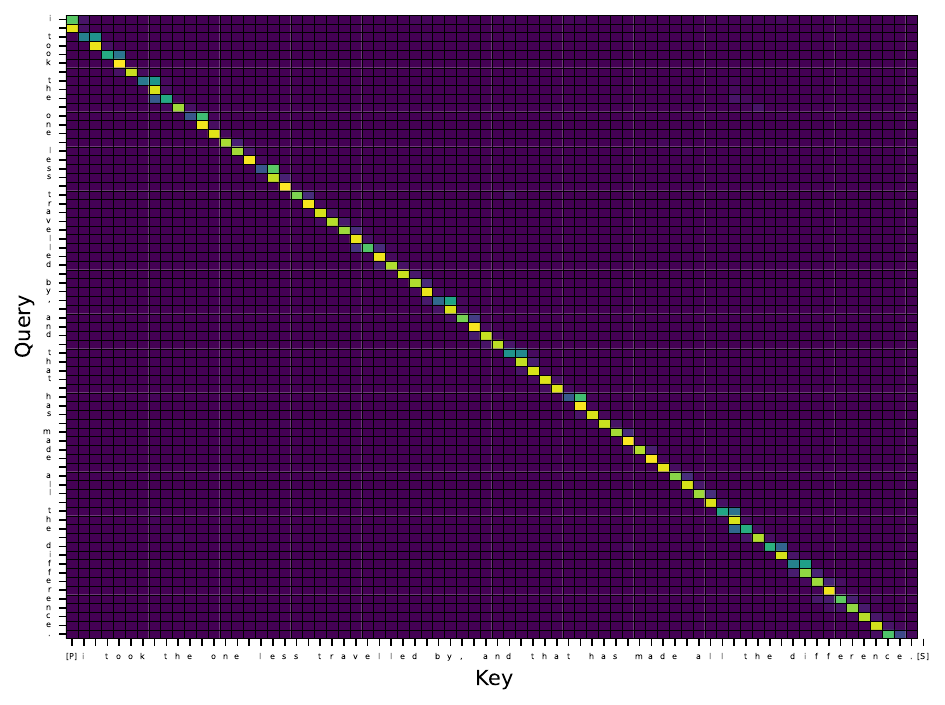}
  % \caption{}
  % \label{}
\end{subfigure} \hfill
\begin{subfigure}{.237\textwidth}
  \includegraphics[width=\textwidth]{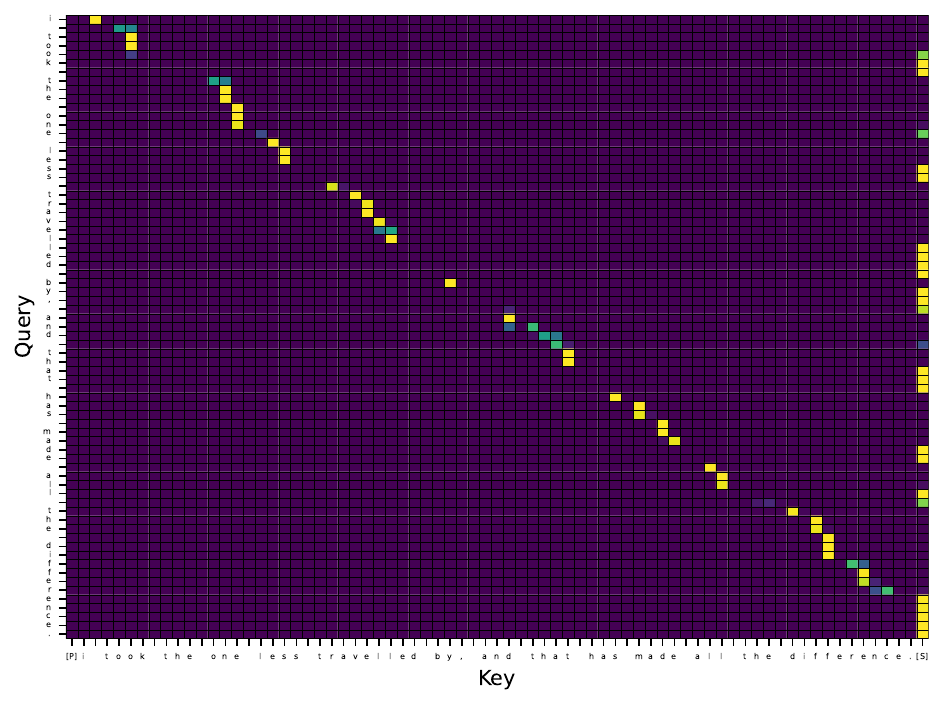}
  % \caption{}
  % \label{}
\end{subfigure}
% layer 3 (lowest layer with NVIB)
\begin{subfigure}{.237\textwidth}
  \includegraphics[width=\textwidth]{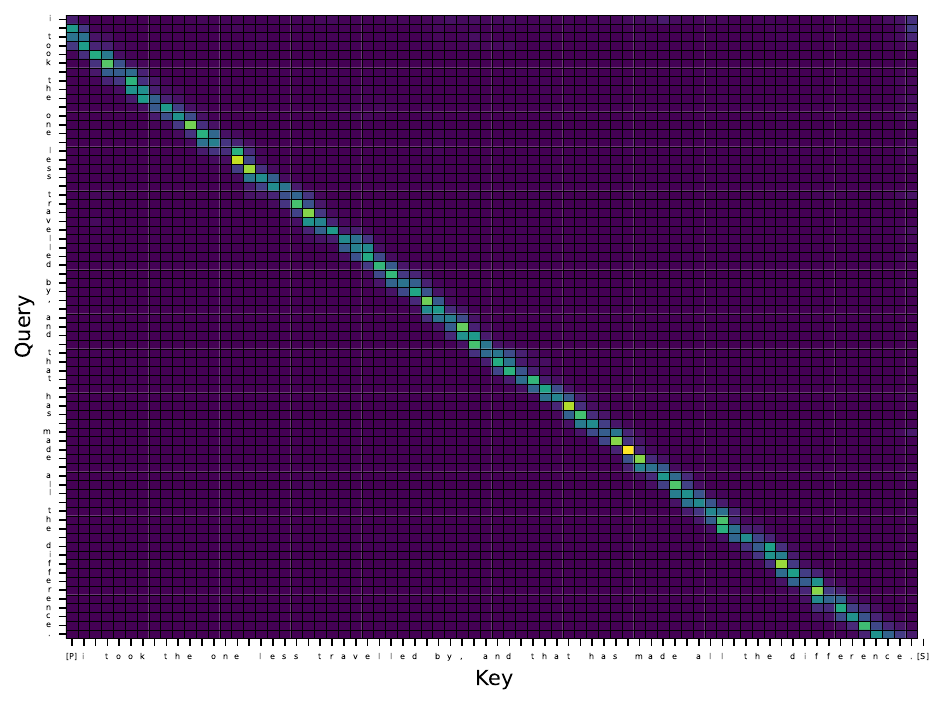}
  % \caption{}
  % \label{}
\end{subfigure} \hfill
\begin{subfigure}{.237\textwidth}
  \includegraphics[width=\textwidth]{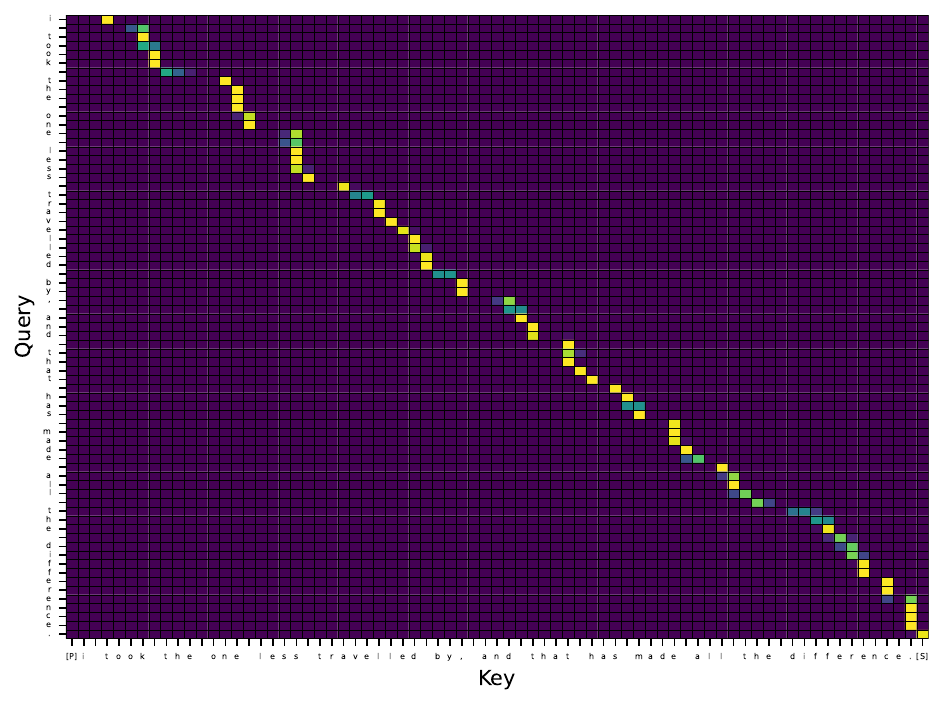}
  % \caption{}
  % \label{}
\end{subfigure}
\caption{Self-attention patterns of the last 3 layers of 6-layer Transformer encoders. \textbf{Left}: Standard self-attention. \textbf{Right}: With NVIB regularisation. \textbf{Sentence}: "I took the one less travelled by, and that made all the difference." Dark purple is 0 and light yellow is 1 for the attention values.}
\label{fig:attention5}
\end{figure}

In \Cref{fig:attention1,fig:attention2,fig:attention4,fig:attention5} we include additional visualisations of the self-attention weights.
%to make sure we are not cherry picking too much

% \begin{figure}[!h]
% % layer 5 (highest layer with NVIB)
% \begin{subfigure}{.237\textwidth}
%   \includegraphics[width=\textwidth]{figures/attention_plots/T6/Attention_3_layer_5.pdf}
%   % \caption{}
%   % \label{}
% \end{subfigure} \hfill
% \begin{subfigure}{.237\textwidth}
%   \includegraphics[width=\textwidth]{figures/attention_plots/T3nvib3/Attention_3_layer_5.pdf}
%   % \caption{}
%   % \label{}
% \end{subfigure}
% % layer 4 (middle of 3 layers with NVIB)
% \begin{subfigure}{.237\textwidth}
%   \includegraphics[width=\textwidth]{figures/attention_plots/T6/Attention_3_layer_4.pdf}
%   % \caption{}
%   % \label{}
% \end{subfigure} \hfill
% \begin{subfigure}{.237\textwidth}
%   \includegraphics[width=\textwidth]{figures/attention_plots/T3nvib3/Attention_3_layer_4.pdf}
%   % \caption{}
%   % \label{}
% \end{subfigure}
% % layer 3 (lowest layer with NVIB)
% \begin{subfigure}{.237\textwidth}
%   \includegraphics[width=\textwidth]{figures/attention_plots/T6/Attention_3_layer_3.pdf}
%   % \caption{}
%   % \label{}
% \end{subfigure} \hfill
% \begin{subfigure}{.237\textwidth}
%   \includegraphics[width=\textwidth]{figures/attention_plots/T3nvib3/Attention_3_layer_3.pdf}
%   % \caption{}
%   % \label{}
% \end{subfigure}
% \caption{\textbf{Left}: Transformer self-attention and \textbf{Right}: Transformer with NVIB with Denoising self-attention in the final 3 layers.}
% \label{fig:attention3}
% \end{figure}

\section{Probing Classifiers} \label{apx: probingclassifiers}
We used two types of probing classifiers to perform our tasks. First, 
we employ an attention-based probing classifier to operate on top of the set of representations for predicting the specified property. This would be similar to having a learnable [CLS] token which is used as the representation of the sequence in BERT-like models. This is also in line with the findings of \citet{pimentel-etal-2022-attentional} that the way we do the probing should resemble how the model itself would use the information within its architecture.  In particular, we first map the representations into a new space with a 2 layer MLP. Then, we compute the attention with a learnable query vector. Finally, we linearly project the resulting vector into the number of classes for each task. We refer to this probe as \textit{Attention-based probe}. 
Second, we tried a less complicated and more common type of probing in which we first aggregate the set of representation vectors by mean and then apply a 2 layer MLP with ReLU non-linearity to perform the task. We refer to this probe as \textit{Aggregating probe}.

\section{SentEval Tasks}
\label{apx:senteval}
\subsection{Model}
We employ the Aggregating probe for performing this task. We froze our models and trained the probes for 10 epochs with a batch-size of 128. The hidden dimension for the probe is set to 256. We trained the model with Adam optimizer with a learning rate of $1e-4$. We report the test set accuracy for the best-performing model in terms of validation accuracy. 

\subsection{Supplementary Results}
We report the results in \Cref{tab:senteval} on a subset of 7 of the 10 SentEval tasks as sentence length (\textbf{SentLen}), word content (\textbf{WC}) and semantic odd man out (\textbf{SOMO}) tasks are too challenging for our models when encoding from a character level.
\begin{table*}[t]
\small
\centering
\begin{tabular}{l|llllllll}
\toprule
& Layer & \textbf{CoordInv} & \textbf{ObjNum} & \textbf{TreeDepth} & \textbf{TopConst} & \textbf{BShift} & \textbf{Tense}  & \textbf{SubjNum}  \\ \midrule
Chance   &       & $0.5$      & $0.5$    & $0.125$     & $0.05$     & $0.5$    & $0.5$    & $0.5$ \\ \midrule
& 1 & $0.5023$ & $0.6498$ & $0.2200$ & $0.2880$ & $0.5006$ & $0.7306$ & $0.6500$ \\
& 2 & $0.5144$ & $0.7255$ & $0.2350$ & $0.3724$ & $0.4994$ & $0.7891$ & $0.7131$ \\
& 3 & $0.5190$ & $0.7547$ & $0.2594$ & $0.4261$ & $0.5055$ & $0.8263$ & $0.7297$ \\
Transformer & 4 & $0.5196$   & $0.7687$ & $0.2692$    & $0.4368$   & $0.5108$ & $0.8114$ & $0.7545$ \\
& 5 & $0.5196$   & $0.7737$ & $0.2736$    & $0.4369$   & $0.5304$ & $0.8320$  & $0.7435$  \\
& 6 & $0.5227$   & $0.7756$ & $0.2736$    & $0.4212$   & $0.5465$ & $0.8384$ & $0.7683$ \\ 
\midrule
& 1 & $0.5037$ & $0.7646$ & $0.2349$ & $0.3323$ & $0.5007$ & $0.8344$ & $0.7285$ \\
& 2 & $0.5069$ & $0.7859$ & $0.2511$ & $0.4243$ & $0.5108$ & $0.8379$ & $0.7777$ \\
& 3 & $0.5110$ & $0.7963$ & $0.2589$ & $0.4453$ & $0.5466$ & $0.8606$ & $0.7844$ \\
NVIB & 4& $0.5111$   & $0.7879$ & $0.2655$    & $0.5290$   & $0.5361$ & $0.8481$ & $0.7943$\\
& 5 & $0.5299$   & $0.7660$ & $0.2651$    & $0.5283$    & $0.5571$ & $0.8371$ & $0.7793$  \\
& 6 & $\textbf{0.5523}$    & $\textbf{0.8207}$ & $\textbf{0.2923}$    & $\textbf{0.5766}$   & $\textbf{0.6075}$ & $\textbf{0.8531}$   & $\textbf{0.8038}$\\ 
\bottomrule
\end{tabular}
\caption{Performance on Senteval tasks.}
\label{tab:senteval}
\end{table*}

% Relative difference - between layers each line is a task x is layers y is F1 score.
% Tasks coordinv, bigram shift, top const.
% Different markers and colours and lines
% Purple green Yellow

\section{Arxiv Classification Task}
\label{apx:arxiv}
Our goal here is to compare the representations and not have ultimate performance in the task, thus we do not fine-tune the models. Hence, we only evaluated our models on the large division of the task, i.e., ArXiv-L which consist of 1000 samples for each sub-area leading to 20000 samples in total. 
We employ the Attention-based probe to perform this task as it is quite a challenging task which requires the information in the vectors to be better managed by the Attention mechanism and also more similar to the way the model itself would perform the task. 
The hidden dimension of the MLP is set to 256 and  the query, key, and value matrices are set to the same dimension as the model dimension, namely, 512. We train the classifier with a batch size of 256 for 50 epochs with Adam optimizer with a learning rate of $1e-3$. Following \citet{hofmann-etal-2022-embarrassingly} we report the test F1 for the best-performing model in terms of validation F1.  %For the Arxiv-L dataset each subarea has 1000 samples which would be 20000 samples in total. 

\section{Quantification of Word Resemblance}
\label{apx:f1}
We observe a strong resemblance between words and the vertical bands in the final-layer Attention maps of the NVIB integrated model. Therefore, we quantify this similarity as follows. First, we take the $argmax$ over the Key dimension of an Attention map and extract the contiguous segments from the resulting vector.  Then, we compute the intersection between the set of obtained segments and the set of words in the sequence. In particular, for each segment and word, we compute the number of intersecting characters (i.e., the length of the longest common substring) as a measure of their overlap. This would lead to a rectangular matrix of scores. Then, we perform the Hungarian matching algorithm \cite{kuhn1955hungarian} to find the best 1-1 match between the two sets. Afterward, for each matched word and segment, we compute Precision ($P$), Recall ($R$), and $F1$ measure as
\begin{equation}
    P = \frac{\text{longest common substring length}}{\text{segment length}}
\end{equation}
and 
\begin{equation}
    R = \frac{\text{longest common substring length}}{\text{word length}}.
\end{equation}

We reported the average macro $F1$, $P$, $R$ over the validation set of our training data. For the baseline Transformer, as it usually predicts units of length one or two which are within a single word the $P$ would be high as opposed to its $R$ value. 
\end{document}